\documentclass[10pt,twocolumn,letterpaper]{article}

\usepackage[pagenumbers]{cvpr} %

\usepackage{amsmath,amsfonts,bm}

\def\eqref#1{equation~\ref{#1}}

\def\1{\bm{1}}

\def\rvepsilon{{\mathbf{\epsilon}}}

\def\rvz{{\mathbf{z}}}

\def\rmZ{{\mathbf{Z}}}

\def\vp{{\bm{p}}}

\def\vv{{\bm{v}}}

\def\mI{{\bm{I}}}

\def\mO{{\bm{O}}}
\def\mP{{\bm{P}}}

\def\mX{{\bm{X}}}

\DeclareMathAlphabet{\mathsfit}{\encodingdefault}{\sfdefault}{m}{sl}
\SetMathAlphabet{\mathsfit}{bold}{\encodingdefault}{\sfdefault}{bx}{n}

\def\sP{{\mathbb{P}}}

\def\sW{{\mathbb{W}}}

\newcommand{\original}[1]{#1^\downarrow}
\newcommand{\extend}[1]{{#1}}
\newcommand{\latent}{{\rmZ}}
\newcommand{\elatent}{\extend{\rmZ}}
\newcommand{\lat}{{\rvz}}

\newcommand{\fvec}{{\vv}}
\newcommand{\efvec}{{\extend{\fvec}}}
\newcommand{\oefvec}{\extend{\hat{\fvec}}}

\newcommand{\pos}{\vp}
\newcommand{\noise}{\rvepsilon}

\newcommand{\occ}{\extend{\mO}}
\newcommand{\sliding}{\sW}
\newcommand{\patch}{\phi}

\newcommand{\pc}{\sP}
\newcommand{\image}{\mathcal{I}}
\newcommand{\camera}{\mP}

\newcommand{\SLat}{\textsc{SLat}}

\newcommand{\first}{\cellcolor{red!30}}
\newcommand{\second}{\cellcolor{orange!30}}
\newcommand{\third}{\cellcolor{yellow!30}}

\newcommand*\circled[1]{\tikz[baseline=(char.base)]{
            \node[shape=circle,draw,inner sep=0.3pt] (char) {#1};}}

\usepackage{kotex}
\usepackage{cuted}
\usepackage{subcaption}
\usepackage[export]{adjustbox} %
\usepackage{mathtools}
\usepackage{booktabs}
\usepackage[table]{xcolor}
\usepackage{tikz}

\usepackage{float}
\usepackage{url}

\usepackage{algorithm}
\usepackage{algpseudocode}
\usepackage{algorithmicx}
\usepackage{kotex}
\usepackage{comment}
\usepackage{makecell}
\usepackage[accsupp]{axessibility}

\definecolor{cvprblue}{rgb}{0.21,0.49,0.74}
\usepackage[pagebackref,breaklinks,colorlinks,allcolors=cvprblue]{hyperref}

\def\paperID{} %
\def\confName{CVPR}
\def\confYear{2026}

\title{Extend3D: Town-Scale 3D Generation}

\author{Seungwoo Yoon
\quad
Jinmo Kim
\quad
Jaesik Park\footnotemark[1]
\\
Seoul National University
\\
{\tt\small \{dotori000, jmkim1012, jaesik.park\}@snu.ac.kr}
\\
}

\begin{document}
\twocolumn[{%
\maketitle
\vspace{-10mm}

\begin{center}
  \includegraphics[width=\linewidth]{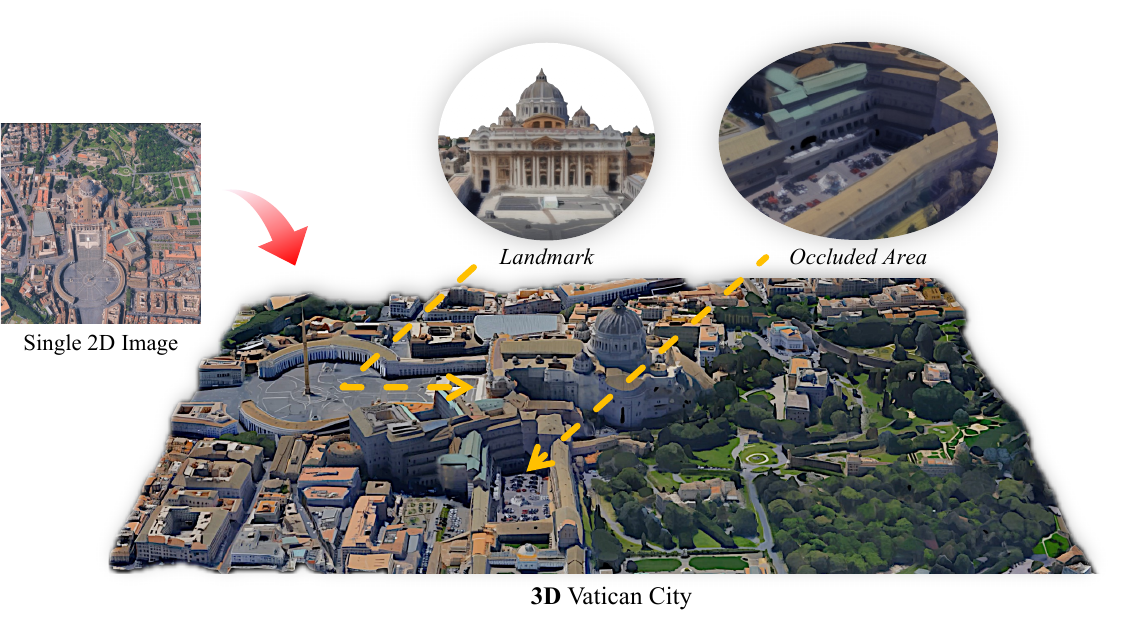}
  \vspace{-7mm}
  \captionof{figure}{\textbf{The result of Extend3D.} We generated a large-scale 3D scene from an image of Vatican City captured from Google Earth~\citep{google-earth}.}
  \label{fig:teaser}
\end{center}
\vspace{4mm}
}]
\begingroup
\renewcommand{\thefootnote}{\fnsymbol{footnote}}
\footnotetext[1]{Corresponding author.}
\endgroup

\begin{abstract}

In this paper, we propose Extend3D, a training-free pipeline for 3D scene generation from a single image, built upon an object-centric 3D generative model.
To overcome the limitations of fixed-size latent spaces in object-centric models for representing wide scenes, we extend the latent space in the $x$ and $y$ directions. 
Then, by dividing the extended latent space into overlapping patches, we apply the object-centric 3D generative model to each patch and couple them at each time step.
Since patch-wise 3D generation with image conditioning requires strict spatial alignment between image and latent patches, we initialize the scene using a point cloud prior from a monocular depth estimator and iteratively refine occluded regions through SDEdit.
We discovered that treating the incompleteness of 3D structure as noise during 3D refinement enables 3D completion via a concept, which we term under-noising.
Furthermore, to address the sub-optimality of object-centric models for sub-scene generation, we optimize the extended latent during denoising, ensuring that the denoising trajectories remain consistent with the sub-scene dynamics.
To this end, we introduce 3D-aware optimization objectives for improved geometric structure and texture fidelity.
We demonstrate that our method yields better results than prior methods, as evidenced by human preference and quantitative experiments. 
\href{http://seungwoo-yoon.github.io/extend3d-page}{project page}
\end{abstract}
    
\section{Introduction}
\label{introduction}

In the modern era, 3D scene assets are essential across fields such as game development, filmmaking, animation, simulation, and other areas of content production. 
Creating detailed 3D scenes requires substantial human effort and resources, even with the provided 3D assets. 
Therefore, a tailored generative model for 3D scenes would help reduce such costs and enhance productivity in industries.

Despite recent advances in 3D generative models, which have enabled the creation of production-ready high-quality 3D objects, generating large-scale 3D scenes remains challenging. 
One of the main challenges is that most current 3D datasets~\citep{deitke2023objaverse, abo2022, 3dfuture2021} consist of object-centric data, and lack cases with complex arrangements of multiple objects and a background. 
Consequently, previous data-centric approaches were unable to generate large general scenes. 
Moreover, existing latent generative models~\citep{trellis2025, hunyuan2} represent 3D data with a fixed latent size, thereby limiting the level of detail of generated results. 
As the 3D scene grows in size, the output becomes blurry due to the limited latent dimensionality, resembling a low-resolution image.
To adequately represent the scene's details, the latent size should be adapted to the scale of the result.

Therefore, research has been conducted to develop training-free pipelines for generating 3D scenes using object-centric models. 
Previous work has explored generating 3D scene blocks through an outpainting process~\citep{syncity2025, 3dtown2025}. 
However, results from these approaches indicate that outpainting can degrade block consistency, particularly in large-scale scenes, making seams visible.
Moreover, they rely entirely on the sub-scene generation capabilities of object-centric models, which are not sufficient.

In this paper, we introduce \textbf{Extend3D}, a novel training-free pipeline for generating 3D scenes from a single image. 
To achieve greater detail and scalability in large-scale 3D scene generation, we have expanded the latent space of a pre-trained 3D object generation model. 
Inspired by training-free, high-resolution image generation methods, such as those presented in recent  works~\citep{multidiffusion2023, syncdiffusion2023, scalecrafter2024, demofusion2024, megafusion2025, accdiffusion2024, cutdiffusion2024}, we divide the extended latent space into overlapping patches and generate them simultaneously. 
Unlike previous outpainting methods, our approach automatically refines fine object details within the scene. 
This is possible because neighboring overlapping patches can influence each other, increasing the likelihood of accurately reconstructing their 3D representations.

However, there are challenges in 2D-3D spatial alignment and the object-centrality of pretrained models. 
To overcome this, we used the input image and the point cloud extracted from the monocular depth estimator~\citep{wang2025moge2} as priors to initialize and optimize the extended latents. 
We initialize the structure from the point cloud and refine the occluded regions using SDEdit~\citep{meng2022sdedit} with \textit{under-noising}.
We optimize the latents at each time step using \textit{3D-aware optimization objectives} to align the image and point cloud, ensuring that the denoising paths remain consistent with the sub-scene dynamics.

The qualitative results show that our method is scalable and generalizable. 
Through human preference and quantitative experiments, we demonstrate that our method outperforms state-of-the-art models in terms of geometry, appearance, and completeness, and is more faithful to the given image.
Through an ablation study, we also demonstrate that overlapping patch-wise flow, initialization, and optimization are crucial for training-free 3D scene generation.

The main contributions of this paper are:
\begin{itemize}
    \item We extend the latent space to integrate object-centric models into 3D scene generation, enabling a more generalizable and scalable generation pipeline.
    \item We introduce an overlapping patch-wise flow with image conditioning that captures local information and mitigates errors arising from object-centric models.
    \item We incorporate an iterative under-noised SDEdit process and 3D-aware optimization to complete occluded regions in the monocular depth point cloud and to overcome the deviation of object-centric models from scene dynamics.
\end{itemize}

\section{Related Work}
\label{related_work}

\textbf{3D generative models.} 
There have been numerous recent studies on generative models that can generate 3D objects conditioned on text or images. 
Currently, their main approach is the latent flow model~\citep{flow2022, rombach2022high} applied to voxel-based or set-based latents. 

Trellis~\citep{trellis2025} generates 3D Gaussians~\citep{3Dgaussians2023}, radiance field~\citep{nerf2020}, and mesh, using two steps of latent flow models where each generates a voxelized sparse structure and structured latents. 
Hunyuan3D~\citep{hunyuan2} utilizes the latent flow model to generate shapes with set-based latents, as proposed in \cite{vecset}. 
TripoSG~\citep{triposg} also uses the set-based latent representation of \cite{vecset} to generate a mesh.
These models have the limitation that they are trained with object-centric datasets. 
Moreover, structurally, current flow-based approaches suffer from the limitation that their latent size is predefined, so the output 3D can only have a confined range of details. 
We solve these problems by extending the latents to represent a large-scale scene.

To overcome the issues of object-centric models, some attempts have been made to train models using 3D scene datasets. BlockFusion~\citep{blockfusion} trains a diffusion model to generate cropped sub-scenes and generate the scene by extrapolation. PDD~\citep{pyramid} trains a multi-scale diffusion model for coarse-to-fine scene generation. 
LT3SD~\citep{meng2025lt3sd} generates a 3D scene hierarchically with a latent tree representation. 
NuiScene~\citep{nuiscene} trains an autoregressive model with chunk VAE and vector sets.
Nevertheless, since all of these methods are trained on limited datasets, they can generate 3D scenes with fewer categories than object-centric models. 
They also do not consider detailed model conditioning, such as image conditions, when designing hierarchical frameworks. 
Unlike them, our method can generate general 3D scenes with detailed image conditioning.

\vspace{2mm}
\noindent\textbf{Training-free 3D scene generation.} 
Recent advances in object-centric 3D generative models and the shortage of 3D scene datasets have led researchers to develop training-free 3D scene generation pipelines using these object-centric models.

SynCity~\citep{syncity2025} generates tiles of 3D sub-scenes sequentially with Trellis from a text using Flux inpainting~\citep{flux2024}. 
Because SynCity attaches separate 3D sub-scenes, there are inconsistencies between tiles, and seams are visible. 
An image-to-3D scene generation pipeline, 3DTown~\citep{3dtown2025}, initializes a scene with the point cloud from VGGT~\citep{wang2025vggt} and then completes it patch by patch using RePaint~\citep{lugmayr2022repaint} and Trellis. 
Although 3DTown can generate 3D towns from images with high fidelity, it can only be used with restricted input due to the limitations of object-centric models (\textit{e.g.}, vanishing floors). 
Also, regardless of initialization, some objects in the scene ignore certain input information, such as rotation. EvoScene~\citep{evoscene} further leverages a video diffusion model~\citep{wan2025} on 3DTown, but suffers from similar problems.

To address the problems of separate and sequential 3D sub-scene generation, we simultaneously generate 3D sub-scenes with interacting denoising paths.
With small transitions between overlapping patches, the generation process can effectively capture local information and prevent geometrical errors through simultaneous generation. 
Also, unlike previous works that rely solely on sub-scene generation using an object-centric model, we optimize the latent representation at each step to prevent paths from transitioning from sub-scene to object dynamics.

\vspace{2mm}
\noindent\textbf{Training-free high-resolution image generation.}
In the field of image generation, training-free high-resolution image generation has been widely researched and has led to massive discoveries on the dynamics of the scaled-up latent denoising process. The primary purpose of this area is to generate high-resolution images from pre-trained models trained on relatively low-resolution data. 

MultiDiffusion~\citep{multidiffusion2023} generates a high-resolution image from text with an extended 2D latent with overlapping patches. 
DemoFusion~\citep{demofusion2024} solves the object repetition problem of Multidiffusion with two ideas: progressive upsampling and dilated sampling. 
Later research~\citep{cutdiffusion2024, accdiffusion2024}, additionally refines dilated sampling. 

When these methods are naively applied to extended 3D latent generation, however, we found that they fail to generate 3D scenes with high fidelity due to the unique dynamics of the model's image-condition, 3D, and object centrality.
For instance, the floor vanishes, or poorly correlated patches lead to repeated objects.
We therefore provide structure priors to generate a high-fidelity 3D scene.

\vspace{2mm}
\noindent\textbf{Generation with priors.} 
Several studies are trying to provide priors for pre-trained generative models for various purposes. 
SDEdit~\citep{meng2022sdedit} is a representative method of image editing that can be applied to \cite{ddpm2020, song2021denoising, rombach2022high, flow2022}. 
SDEdit partially noise the original image, producing an edited image whose perturbed distributions retain the original image's style, meeting the intended image style.
Readout Guidance~\citep{readout2024} trains a small neural network to extract properties (\textit{e.g.}, pose, depth, or edges) from the intermediate latent representation. 
Then, it computes the loss with respect to the property and provides a loss gradient as guidance, similar to the classifier guidance~\citep{dhariwal2021diffusion}. 

We apply SDEdit in our Extend3D to refine the initialized structure. 
Unlike image editing, we propose an under-noising technique designed for the 3D completion task. 
Also, instead of guidance, we optimize the intermediate latent with a loss explicitly designed for 3D scene generation, assuming that the priors have ground-truth knowledge of 3D structure and texture. 

\begin{figure*}
    \centering
    \includegraphics[width=0.9\linewidth]{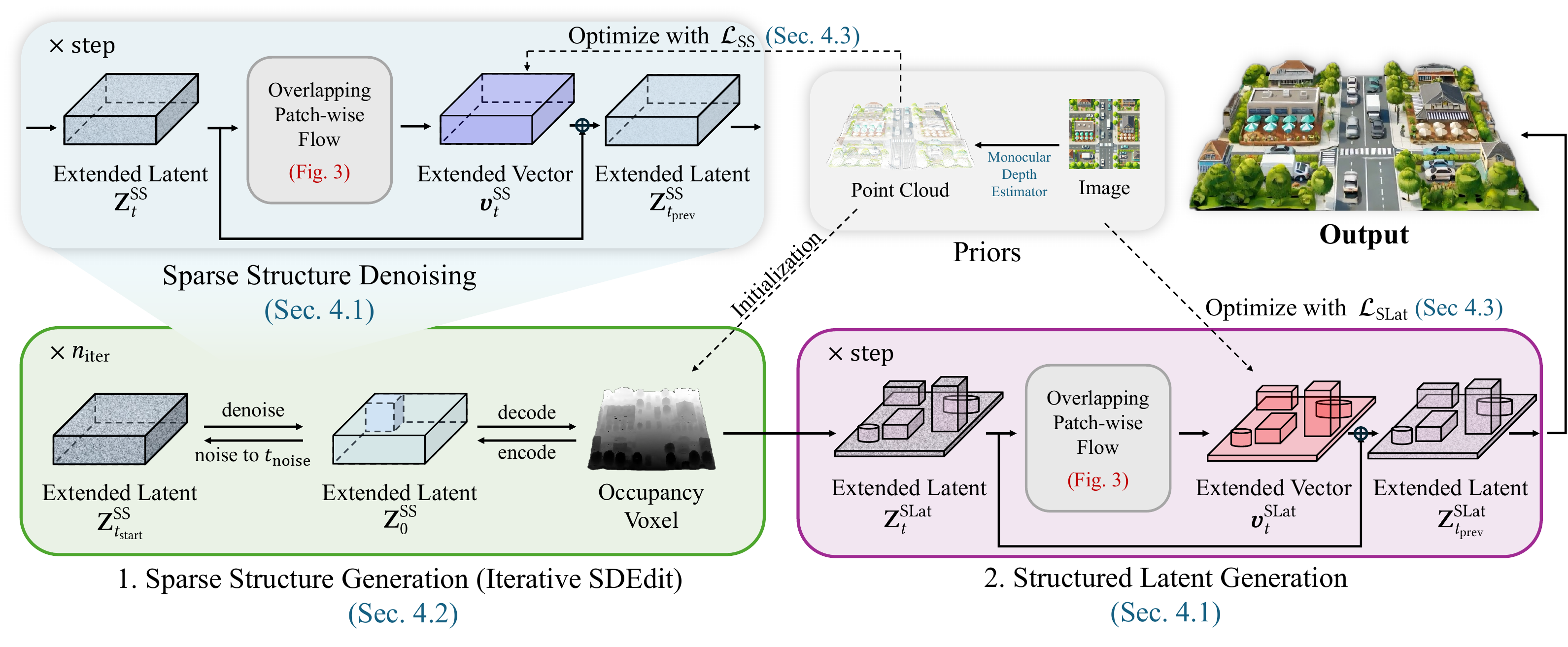}
    \vspace{-3mm}
    \caption{\textbf{An overall pipeline of our Extend3D.} Extend3D consists of two parts: sparse structure generation and structured latent generation. In the denoising part of both steps, an overlapping patch-wise flow was used (\cref{Overlapping Patch-wise Flow} and \cref{fig:patch}). In sparse structure generation, iterative SDEdit is used to initialize the structure (\cref{Initialize with Prior}). Vector fields in both steps are optimized with priors (\cref{Optimize with Prior}).}
    \label{fig:pipeline}
    \vspace{-3mm}
\end{figure*}

\section{Preliminaries}

\subsection{Latent Flow Model for 3D Generation}
\label{sec:pretrained_preliminary}
A modern approach for high-quality 3D generative models is the latent flow model. 
They use voxelized latents of fixed size or set-based latents (\textit{e.g.}, point clouds) within a confined region to represent 3D space. 
While our approach is not restricted to a specific generative model, it can be applied to general voxel-based latents or set-based latent flow models. We illustrate our idea using Trellis~\citep{trellis2025}, which is one of the leading 3D generative models.

Trellis generates 3D representations with two steps of latent flow models given a condition $C_\image$ encoded from an image $\image$ by DINOv2~\citep{dinov22024}, and both steps are generalizable to flow models for voxelized or set-based latents.
The first step of the model \textbf{\emph{generates a sparse structure}} (SS) $\{\pos_i\}\subset[M]^3$ (where $[M]:=\{0, 1, ..., M-1\}$), which represents a set of occupied coordinates in a voxel grid.
In sparse structure generation, low-resolution voxelized noise $\latent_1^\mathrm{SS}\in\mathbb{R}^{N\times N\times N}$ is denoised to $\latent_0^\mathrm{SS}$ with vector field $\fvec_\mathrm{SS}$, decoded with decoder $\mathcal{D}$, and activated voxel coordinates are collected as:
\begin{gather}
    \latent_1^\mathrm{SS} \sim \mathcal{N}(\bm0, \mI), 
    \quad 
    \frac{d}{dt}\latent_t^\mathrm{SS}=\fvec_\mathrm{SS}(\latent_t^\mathrm{SS}, C_\image, t), \\
    \{\pos_i\}=\{\pos: \mathcal{D}(\latent_0^{SS})_\pos > 0\}.
\end{gather}
As the decoder is trained as a VAE, there is a trained encoder $\mathcal{E}$ that encodes the occupancy grid $O \in \mathbb{R}^{M\times M\times M}$ into a low-resolution latent representation. 
The second step of the model conducts \textbf{\emph{denoising on a structured latent}} (\SLat), where a set-based latent feature is matched to a coordinate of sparse structure as:
\begin{gather}
    \latent_t^\mathrm{\SLat}=\{(\pos_i, \lat_{i, t})\}\subset[M]^3\times\mathbb{R}^l, \\
    \lat_{i, 1} \overset{\mathrm{iid}}{\sim} \mathcal{N}(\bm0, \mI),
    \quad
    \frac{d}{dt}\latent_t^\mathrm{\SLat}=\fvec_\mathrm{\SLat}(\latent_t^\mathrm{\SLat}, C_\image, t),
\end{gather}
with invariant $\pos_i$ and vector field $\fvec_\mathrm{\SLat}$. \SLat\ is then decoded to 3D representations such as 3D Gaussians, radiance field, or mesh by sparse decoders ($\mathcal{D}_{\mathrm{GS}}$, $\mathcal{D}_{\mathrm{NeRF}}$, and $\mathcal{D}_{\mathrm{mesh}}$), and usually. 
In this paper, we will use the notations $\latent_t$ that can refer to both $\latent_t^{\mathrm{SS}}$ and $\latent_t^{\mathrm{\SLat}}$, and $\fvec$ for $\fvec_{\mathrm{SS}}$ and $\fvec_{\mathrm{\SLat}}$ for simplicity.

\subsection{SDEdit}

We introduce SDEdit to refine the initialized structure, treating scene generation as a 3D sub-scene editing task. 
SDEdit noises latent of a ``guide" (\textit{e.g.}, image to be edited) $\latent_0^{(g)}$ to $\latent_{t_\mathrm{start}}$ and denoises it to $\latent_0$ to get the edited result. 
With the added noise, the perturbed distribution meets the intended distribution while preserving information in the guidance. 
Although SDEdit was designed for diffusion models~\citep{score2021}, we can integrate it into flow models, with the following equations:
\begin{gather}
\label{SDEdit_noise}
    \latent_{t_\mathrm{start}}=(1-t_\mathrm{start})\cdot \latent_0^{(g)}+t_\mathrm{start}\cdot\noise;
    \hspace{0.3em}
    \noise\sim\mathcal{N}(\bm0, \mI), \\
    \frac{d}{dt}\latent_t=\fvec(\latent_t, C, t),
\end{gather}
where $C$ refers to the editing condition. 
When $t_\mathrm{start}$ increases, the denoising path gets longer, causing the effect of conditioning and generative models to be enlarged.

\section{Method}
\label{method}

\begin{figure*}
    \centering
    \includegraphics[width=0.9\linewidth]{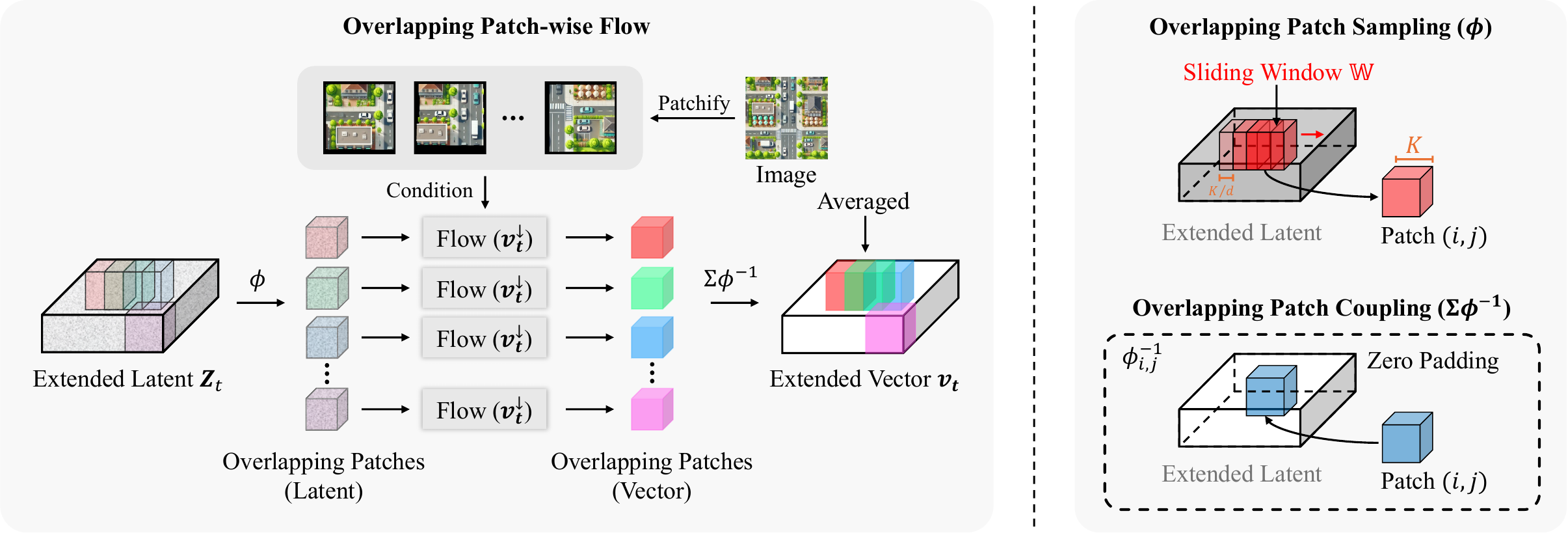}
    \caption{\textbf{Overlapping patch-wise flow.} The extended latent is divided into latent patches with the sliding window. We then obtain the patch vector for each latent patch and merge them into a single extended latent vector, thereby coupling the patches.}
    \label{fig:patch}
    \vspace{-3mm}
\end{figure*}

Extend3D is a training-free pipeline that generates a 3D scene from a single scene image. 
To implement 3D scene generation, we extended the 3D latents of a pre-trained object-centric 3D generative model~\citep{trellis2025} to represent more detailed, larger 3D scenes. 
We extend the latents in the $x$ and $y$ coordinates, and a portion of the extended latent serves as a conventional latent for the pre-trained object-centric 3D generative model.

To handle extended latents, we divide them into overlapping patches, generated simultaneously via separate but coupled denoising paths conditioned on image patches (\cref{Overlapping Patch-wise Flow}). 
Additionally, to address the underlying issues of the object-centric model (\textit{e.g.}, vanishing floor, inability to generate sub-scenes, and randomly rotated objects) and to mitigate the problems associated with patch-wise generation (\textit{e.g.}, repeated objects and seams between patches), we incorporate priors into the generation process. 
We first initialize the scene with a point cloud from a depth estimator and perform iterative under-noised SDEdit.
This completes the occluded area and refines the scene while generating the structure (\cref{Initialize with Prior}). 
We then optimize the scene at every time step using the point cloud and an image of the entire scene. 
We also propose a loss function that treats the point cloud as a prior for the voxel-based latent (\cref{Optimize with Prior}). 
The overall pipeline is illustrated in \cref{fig:pipeline} and \cref{Algorithms}.

\subsection{Overlapping Patch-wise Flow}
\label{Overlapping Patch-wise Flow}

In order to generate a detailed 3D structure and texture, we introduce an extended latent for sparse structure $\elatent_t^\mathrm{SS}\in\mathbb{R}^{aN\times bN\times N}$ and 
an extended \SLat\
$\elatent_t^\mathrm{\SLat}\subset[aM]\times[bM]\times[M]\times \mathbb{R}^l$ where $\elatent_t$ can refer to both. 
Here, $a$ and $b$ are extension factors.
(From here, we will use $\downarrow$ to notate non-extended latents or vectors.)  

We divide these latents into overlapping patches with a division factor $d$. 
We refer to the $(i, j)$-th latent patches as $\patch_{i,j}^\mathrm{SS}(\elatent_t^\mathrm{SS})$ and $\patch_{i,j}^\mathrm{\SLat}(\elatent_t^\mathrm{\SLat})$.
This process can be described as a $N^3$ or $M^3$-sized sliding window $\sliding$ moving with stride $N/d$ or $M/d$ to sample patches, illustrated as sampling in \cref{fig:patch}. 
The patches can be mapped back to their original positions by setting the values at the other positions to zero (zero padding), thereby coupling them, as illustrated in \cref{fig:patch}. 
We represent these inverse mappings as $(\patch_{i,j}^\mathrm{SS})^{-1}$ and $(\patch_{i,j}^\mathrm{\SLat})^{-1}$.
We leave the rigorous definitions of the mappings in \cref{math}.

We also patchify the image condition with $\psi_{i,j}$, which crops the image region to exactly match the $(i,j)$-th 3D patch (see details in \cref{Image Patchification}). 
Similar to MultiDiffusion, we get the vector field of the extended latents by merging the vector fields for each patch, where the overlapping regions are averaged across the patches, as illustrated in the left side of \cref{fig:patch}.
The entire overlapping patch sampling, merging, and denoising process can be formulated as:
\begin{gather}
\label{indep vector}
    \fvec_{i, j}(\elatent_t,\, \image,\, t)=\original{\fvec}\bigl(
      \patch_{i,j}(\elatent_t),\,
      C_{\psi_{i,j}(\image)},\,
      t
    \bigr), \\
\label{extended vector}
\efvec(\elatent_t,\, \image,\, t)
= 
\sum_{i,j}
  \patch_{i,j}^{-1}\!\bigl(
  \fvec_{i, j}
  \bigr)
  ~\oslash~
  \sum_{i,j} \mathbf{1}_{\sliding_{i,j}}, \\
\label{extended_dynamics}
    \frac{d}{dt}\elatent_t=\efvec(\elatent_t, \image,t),
\end{gather}
where $\oslash$ is an element-wise division.
\Cref{indep vector} can be calculated independently from the other patches and in parallel, but the dynamics of different patches, even far away, can be coupled by overlaps.

The advantage of divided but coupled dynamics is the ability to refine errors in other patches. 
By detecting slight movement of the sliding window, our method can identify local information from changes in the image and in latent features between patches. 
Additionally, because some objects are at the centers of patches, we can leverage the object-centric model more effectively. 
The beneficial effect of overlapping patch-wise flow can be found in \cref{fig:ablation}~(A).

Noted in DemoFusion~\citep{demofusion2024}, AccDiffusion~\citep{accdiffusion2024}, and CutDiffusion~\citep{cutdiffusion2024}, dilated sampling is crucial for generating a consistent global structure. 
We apply dilated sampling during the sparse structure generation phase and leave the details to \cref{Dilated Sampling}.

\subsection{Initialize with Prior}
\label{Initialize with Prior}
\begin{figure}[b]
\vspace{-4mm}
\centering
\includegraphics[width=\linewidth]{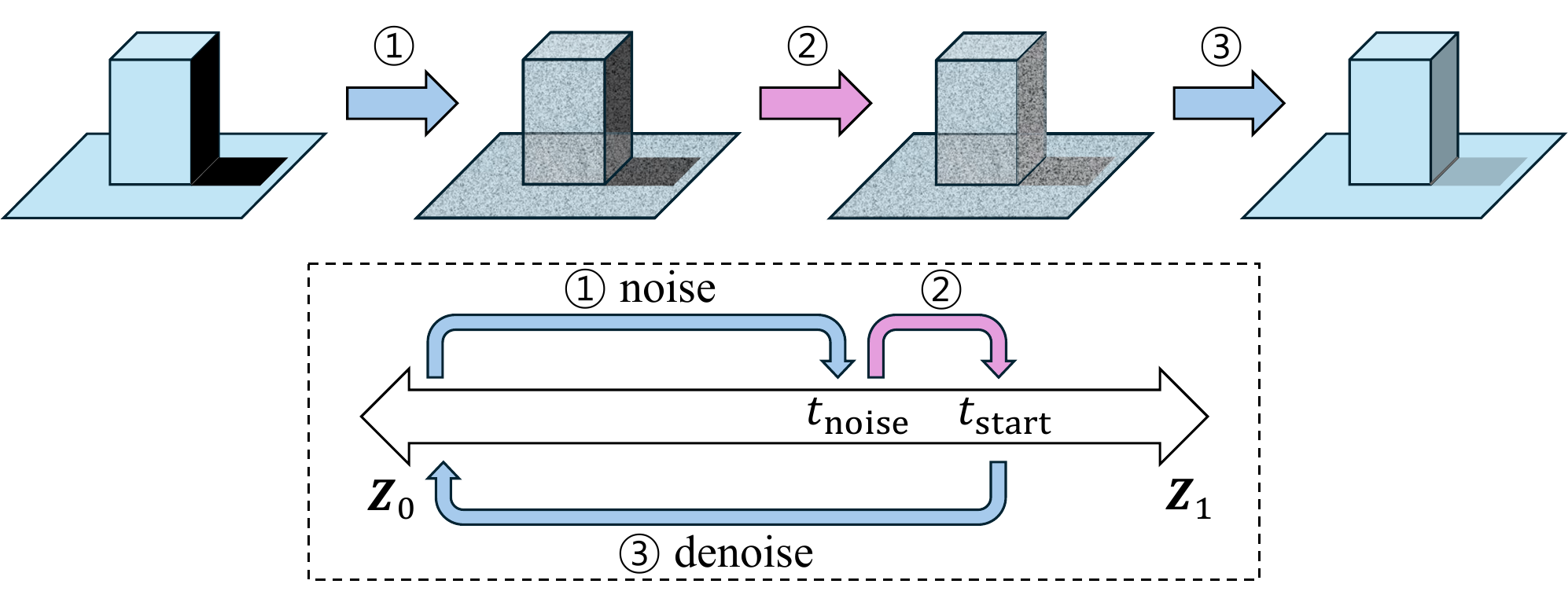}
\vspace{-3.5mm}
\caption{\textbf{Motivation of under-noising}. The blue arrows represent actual noising or denoising, while the purple arrow illustrates how the model is presumed to perceive.}
\label{fig:under_noise}
\end{figure}

When directly denoising sparse structure from pure Gaussian noise using \cref{extended_dynamics}, all patches fail to initialize each sub-scene due to the inherent limitation of the object-centric models. 
Moreover, the coarse structure is determined during the early denoising stage~\citep{megafusion2025}, before the patches are sufficiently coupled, so that the image condition and the 3D latent are not well spatially aligned.
Consequently, the output becomes noisy, fragmented, and unstable as in \cref{fig:ablation} (B). 
This motivates the need for a robust structural prior at initialization.

Inspired by 3DTown~\citep{3dtown2025}, we initialize the scene structure with a point cloud $\pc$ extracted from a monocular depth estimator. 
Specifically, we adopt MoGe-2~\citep{wang2025moge, wang2025moge2} for our Extend3D. 
The predicted point cloud is voxelized into an occupancy grid $\occ_0\in\mathbb{R}^{aM\times bM\times M}$. 
Because the monocular depth estimator cannot infer the occluded regions, the resulting occupancy voxel grid contains empty areas that should be rectified using the pre-trained generative model.  
To address this, with an encoded voxel grid $\elatent_0^{(g)}=\mathcal{E}(\occ_0)$, our Extend3D performs SDEdit. 
Unlike standard SDEdit, which applied \cref{SDEdit_noise}, we introduce \textit{under-noising}:
\begin{equation}
\label{extend3d_noise}
    \elatent_{t_\mathrm{start}}=(1-t_\mathrm{noise})\cdot \elatent_0^{(g)}+t_\mathrm{noise}\cdot\noise,
    \quad
    \noise\sim\mathcal{N}(\bm0, \mI),
\end{equation}
where $t_\mathrm{start}>t_\mathrm{noise}$, ensuring that the latent is denoised more aggressively than it was originally noised. By under-noising the guide structure, the pre-trained model may treat missing or occluded parts as additional noise, illustrated as the arrow \circled{2} in \cref{fig:under_noise}. Finally, the denoising process, represented as arrow \circled{3}, allows such areas to be filled.
This is similar to adding high-frequency noise to enhance image detail in image super-resolution~\citep{LSRNA2025}. We empirically validate this choice in \cref{ablation}.

SDEdit can fill the unwanted empty areas. However, it often fails to fully complete the scene, leaving some holes. 
To mitigate this, as a single SDEdit process partially refines the structure, we apply SDEdit iteratively as: $\occ_n=\mathrm{SDEdit}(\occ_{n-1})$, represented in \cref{fig:pipeline}.
This process iteratively fills the occluded region of $\occ_0$ and eventually completes the scene.

\subsection{Optimize with Prior}
\label{Optimize with Prior}

During denoising, sub-scenes deviate from a scene-like structure toward an object-like structure due to the object-centric model's properties, leading to distortion or a vanishing floor, even with proper initialization. 
To prevent deviation and to align the denoising paths with the conditioning, we optimize the extended latents over time steps using the point cloud and the image. 
When solving \cref{extended_dynamics} with the discrete ODE solver, instead of moving directly along $\efvec(\elatent_t, \image, t)$, we use $\oefvec_t$, an optimized vector starting from $\efvec(\elatent_t, \image, t)$. 
By optimization, we can leverage the pre-trained model for the occluded region while optimizing on the seen region, as in Readout Guidance~\citep{readout2024}.
In addition, optimizing the vector field can improve consistency across patches by simultaneously optimizing the entire scene, as in \cite{syncdiffusion2023}.
We introduce two optimization losses, one for sparse structure generation and one for structured latent generation, as explained in \cref{sec:pretrained_preliminary} and illustrated in \cref{fig:pipeline}.

In the \textbf{\emph{sparse structure generation step}}, we define:
\begin{equation}
\label{ss_optim_loss}
    \mathcal{L}_{\mathrm{SS}}=-\frac{1}{|\pc|}\sum_{\pos\in\pc}\log{\sigma((\mathcal{D}(\elatent_t^{\mathrm{SS}}-t\cdot \oefvec_t)})_\pos),
\end{equation}
where $\sigma$ is a sigmoid function. The loss function is designed to enforce that the initialized voxels do not disappear during the denoising process, motivated by binary cross-entropy loss. It gives a positive signal on predicted voxels where points exist. Voxels with dense point clouds will have more weight in the loss. While this loss can be minimized by increasing the number of voxels, combined with the pre-trained model every time step, it merely prevents the desired voxels from disappearing, rather than creating undesired voxels. 
Moreover, for the same reason, it can smoothly connect the point cloud priors and generated voxels, not just by attaching two distinct voxel grids.
With the $\mathcal{L}_\mathrm{SS}$, we optimize $\oefvec_t$ with Adam optimizer~\citep{adam2015}.

In the \textbf{\emph{structured latent generation step}}, we apply the extended rendering loss~\citep{lpips, ssim} as follows:
\begin{gather}
    \hat{\image}=\mathrm{Render}(\mathcal{D}_{\mathrm{GS}}(\elatent_t^{\mathrm{\SLat}}-t\cdot \oefvec), \camera), \\
    \mathcal{L}_{\mathrm{\SLat}}=\mathrm{LPIPS}(\hat{\image}, \image)-\mathrm{SSIM}(\hat{\image}, \image),
\end{gather}
where $\mathrm{Render}$ is a differentiable renderer (such as Gaussian splatting) and $\camera$ is a camera parameter of an image viewpoint provided by the depth estimator. 
This optimizes the entire scene with an image in the original camera view. 
Because an object-centric model often loses details in scene textures, this optimization helps refine them. 
Also, it makes the seams invisible because the boundary is optimized at every time step, ensuring paths are more consistent with each other. 
With $\mathcal{L}_\mathrm{\SLat}$, we also optimize $\oefvec_t$ with Adam.

Please refer to \cref{Algorithms} for the algorithm details.

\section{Experiments}
\label{experiments}

\begin{figure*}
    \vspace{-3mm}
    \centering
    \includegraphics[width=\linewidth]{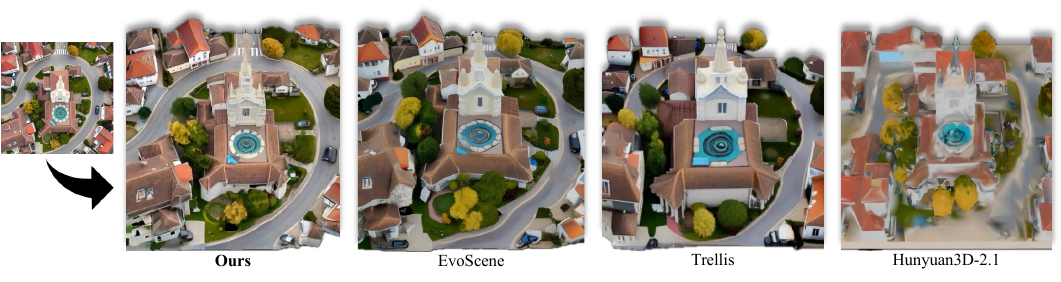}
    \vspace{-7.5mm}
    \caption{\textbf{Qualitative result of our Extend3D.} Our 3D scene generation result (with $a=b=2$) is compared to the results of state-of-the-art 3D generative models. While previous methods may not accurately represent the image or lose scene details, our method effectively expresses the image condition in 3D. The input image is generated using Flux.1 [dev]~\citep{flux2024}. We provide additional results in \cref{More Results}.}
    \label{fig:result}
    \vspace{-1mm}
\end{figure*}

\begin{figure*}[h]
    \centering
    \includegraphics[width=0.75\linewidth]{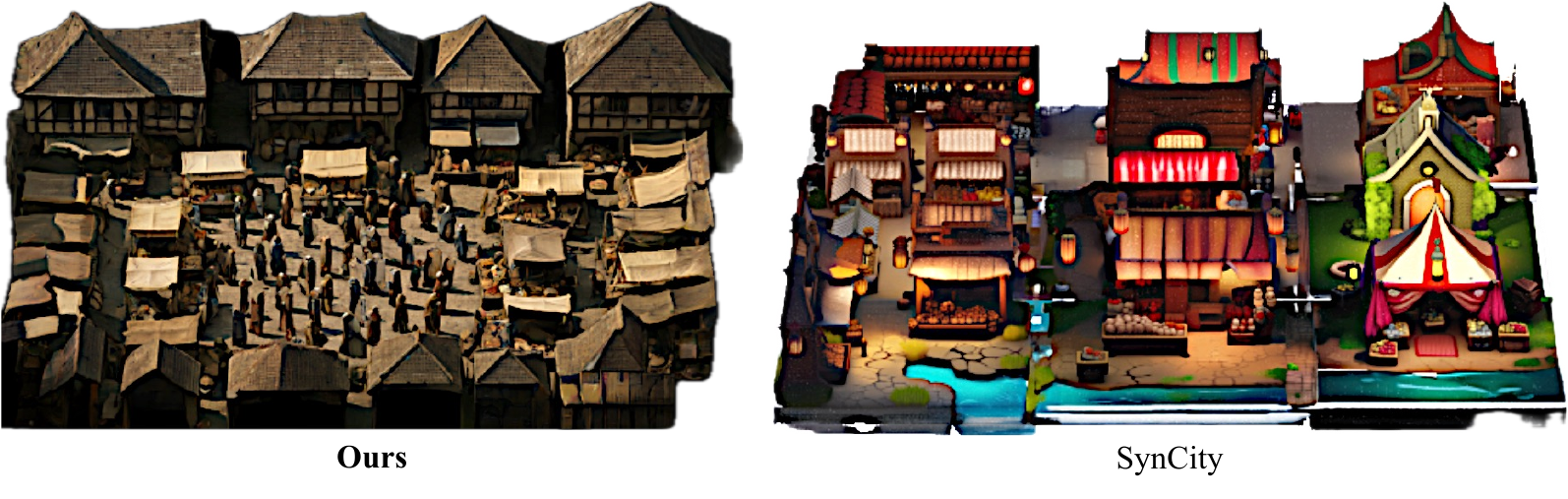}
    \vspace{-3.5mm}
    \caption{\textbf{Qualitative comparison with SynCity.} The results are generated from the text prompt, \textit{medieval market}.}
    \label{fig:syncity}
    \vspace{-3mm}
\end{figure*}

\subsection{Human Preference}
\label{sec:human}

\begin{table}[H]
\vspace{-2mm}
\caption{\textbf{Human preference win rate (\%) of our method.}}
\vspace{-2mm}
\resizebox{\linewidth}{!}{
\begin{tabular}{c|c|c|c|c}
    \toprule
    versus. & Geometry & Faithfulness & Appearance & Completeness \\
    \midrule
    Trellis~\citep{trellis2025} & 50.0 & 66.4 & 67.1 & 62.1 \\
    Hunyuan3D~\citep{hunyuan2} & 73.6 & 75.7 & 75.0 & 75.0 \\
    EvoScene~\citep{evoscene} & 87.1 & 87.9 & 87.1 & 87.1 \\
    \bottomrule
\end{tabular}
}
\label{tab:human}
\vspace{-2mm}
\end{table}

\noindent

To score the visual aestheticity of 3D scenes, we conducted a human preference study. 
We compared our method with Trellis~\citep{trellis2025} and Hunyuan3D-2.1~\citep{hunyuan2}, the current best open-sourced 3D generation models, and EvoScene, which is specifically designed for large-scene generation. 
Human annotators (10 participants) ranked the methods on four criteria: geometry, faithfulness, appearance, and completeness, given 14 images and 3D scenes.
As a result (\cref{tab:human}), our Extend3D outperformed previous methods in four criteria.

\subsection{Quantitative Results}
\label{sec:quantitative}

\begin{table}[h]
\centering
\vspace{-2mm}
\caption{\textbf{Quantitative results.}}
\vspace{-2mm}
\resizebox{\linewidth}{!}{
\begin{tabular}{c|c|c|c|c|c}
    \toprule
    ~ & LPIPS $\downarrow$ & SSIM $\uparrow$ & PSNR $\uparrow$ & CD $\downarrow$ & F-score (0.05) $\uparrow$ \\
    \midrule
    Trellis~\citep{trellis2025} & 0.650 & 0.239 & 10.0 &  0.0315 & 0.442 \\
    Hunyuan3D~\citep{hunyuan2} & 0.683 & 0.255 & 10.4 & 0.0192 & \third 0.567 \\
    EvoScene~\citep{evoscene} & \third 0.482 & \third 0.310 & \third 13.2 & \third 0.0188 & 0.498 \\
    \midrule
    Ours w/o $\mathcal{L}_{\mathrm{\SLat}}$ & \second 0.400 & \second 0.333 & \second 13.8 & \first 0.0078 & \first 0.708\\
    \textbf{Ours} & \first 0.240 & \first 0.611 & \first 20.4 & \second 0.0086 & \second 0.694 \\
    \bottomrule
\end{tabular}
}
\label{tab:quantitative}
\vspace{-2mm}
\end{table}

\noindent
We render the 3D scene into the camera view of the input image using the camera parameter estimator~\citep{wang2025moge2}, and obtain LPIPS, SSIM, and PSNR scores on 100 input images~\citep{chatgpt, flux2024, motionsc, google-earth, urbanscene} spanning diverse wide scenes. 
As shown in \cref{tab:quantitative}, our method achieved the best scores across three metrics, indicating that it is most faithful to the input image in terms of structure and texture.
Using 45 images and ground-truth mesh pairs from the UrbanScene3D dataset~\citep{urbanscene}, we evaluated the geometric results using the Chamfer Distance (CD) and F-score with a threshold of 0.05. 
\Cref{tab:quantitative} shows that our method surpasses the results of the previous methods.

\begin{table}[h]
\centering
\vspace{-2mm}
\caption{\textbf{Comparison between 3D scene generation methods.}}
\vspace{-2mm}
\resizebox{0.7\linewidth}{!}{
\begin{tabular}{c|c|c|c}
    \toprule
    ~ & CLIP $\uparrow$ & HPSv3 $\uparrow$ & Intra-LPIPS $\downarrow$ \\
    \midrule
    SynCity~\citep{syncity2025} & 0.251 & 3.254 & 0.631  \\
    \textbf{Ours} & \textbf{0.276} & \textbf{3.519} & \textbf{0.571} \\
    \bottomrule
\end{tabular}
}
\label{tab:scene_quantitative}
\vspace{-2mm}
\end{table}

We also compared our method with the state-of-the-art training-free 3D scene generation pipeline, SynCity. 
Since SynCity is text-conditioned, whereas ours is image-conditioned, we first generated scene images from keywords using ChatGPT~\citep{chatgpt} and then used Extend3D. 
We render the results of SynCity and Extend3D to get CLIP score~\citep{clip}, HPSv3~\citep{HPSv3}, and Intra-LPIPS~\citep{syncdiffusion2023}. 
Here, Intra-LPIPS refers to LPIPS between patches within a single scene, measuring the patch consistency. 
\Cref{tab:scene_quantitative} shows that our Extend3D is superior to SynCity in text compatibility, quality, and patch-wise consistency. 

\subsection{Qualitative Results}

\Cref{fig:teaser} shows the scalability of our method. 
Given a town-scale scene image, the extended latent can fully capture details, including landmarks and small buildings, and produce a 36$\times$ larger result than the original latent space. 
We present more examples of wide scenes in \cref{fig:large1} and \cref{fig:large2}. %
Also, our Extend3D can represent general scenes.
It can generate a town, a table of foods, a study scene, and an indoor room, illustrated in \cref{fig:result} and \cref{More Results}. In diverse cases, our method outperformed previous 3D generative models. 
Also, compared with SynCity, our method can generate scenes without patch boundaries, as shown in \cref{fig:syncity} and \cref{fig:more8}. 
Moreover, because SynCity generates a scene in an outpainting approach, it cannot refine the unnatural edge of the water.
Compared to EvoScene in \cref{fig:evoscene}, our results had less distorted geometry and more detailed textures.

\subsection{Ablation Study}
\label{ablation}

\begin{figure*}[t]
    \centering
    \includegraphics[width=0.95\linewidth]{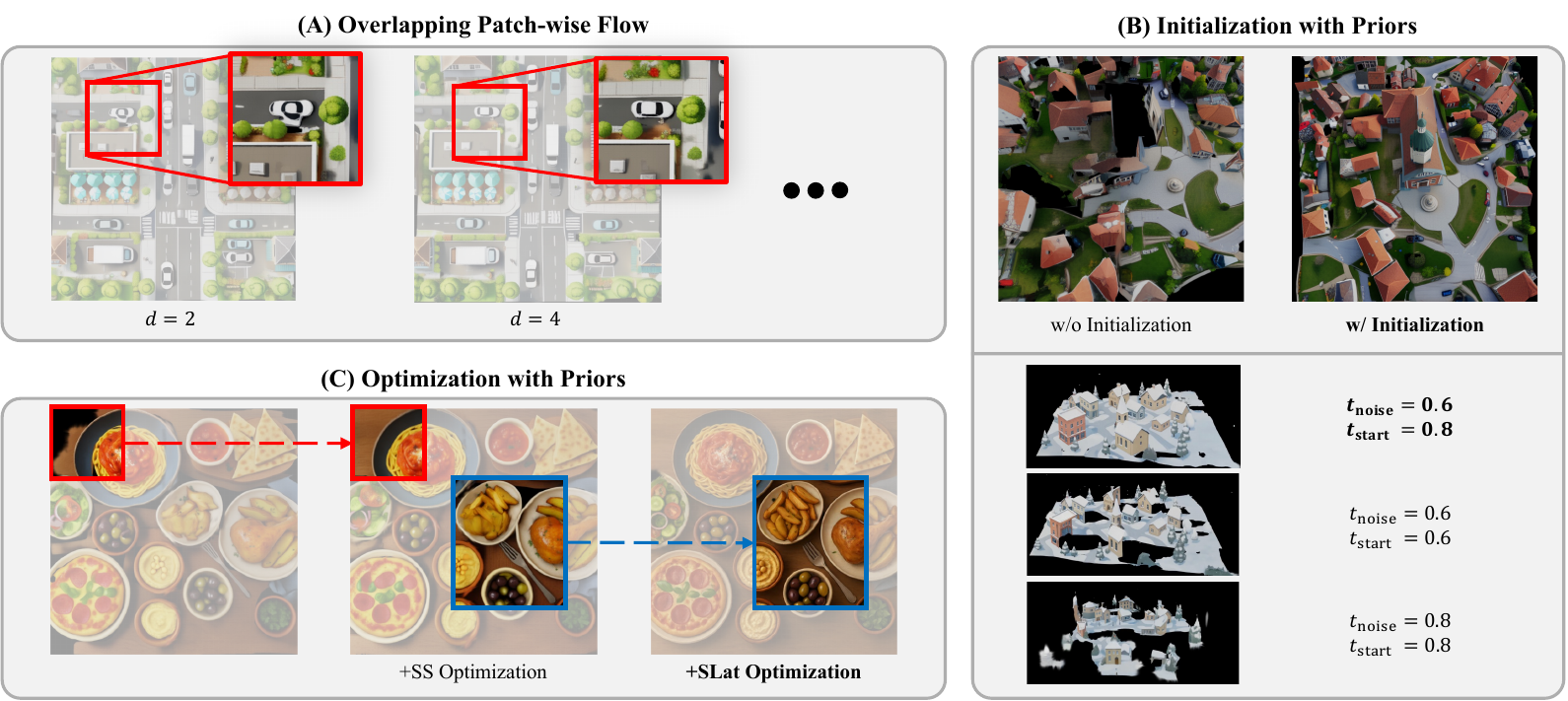}
    \caption{\textbf{Ablation study.} All the images, except for the ablation of under-noising, are taken from the input image camera viewpoint. We set $a=b=2$ to generate the 3D scenes in this figure.}
    \label{fig:ablation}
    \vspace{-3mm}
\end{figure*}

We conducted an ablation study on three proposed methods in our Extend3D and presented the results in \cref{fig:ablation}. When we obtained the results for \cref{fig:ablation} (A) and (B), we did not optimize the latent to emphasize the structural difference.

\vspace{2mm}
\noindent
\textbf{Overlapping Patch-wise Flow.}
We claim that the coupled paths of patches can mutually rectify and effectively capture local information. 
To validate this argument, we compare results for varying division factors. 
As illustrated in \cref{fig:ablation} (A), $d=2$ distorted local structure, while $d=4$ did not. 
These results demonstrate that patch interactions correct each other, and the sliding window's stride enables the extended latent to capture finer details.

\vspace{2mm}
\noindent
\textbf{Initialize with Prior.}
In the first part of \cref{fig:ablation} (B), the results with and without initialization are compared. 
Without initialization (\textit{i.e.}, $t_\mathrm{start}=1$), the structure is totally broken with important objects disappeared, buildings not in proper position, etc. 
We therefore conclude that proper initialization is essential for extended latents. 
In the second part, we compared three results with different $t_\mathrm{noise}$ and $t_\mathrm{start}$. 
When $t_\mathrm{noise}=t_\mathrm{start}$ (usual SDEdit), the structure maintained the holes in the initial point cloud or was destroyed due to the $t_\mathrm{start}$ trade-off in SDEdit.
However, with $t_\mathrm{noise} < t_\mathrm{start}$ (under-noising), the occluded region of the initial structure is completed naturally.

\vspace{2mm}
\noindent
\textbf{Optimize with Prior.} \Cref{fig:ablation} (C) shows the ablation study for optimization with priors. 
Starting from the base model, we sequentially added sparse structure and \SLat\ optimization. 
Without sparse structure optimization, the floor and parts of the objects vanished, as in \cref{More Results}. 
Structured latent optimization could refine seams and distortion between patches compared to those without optimization. 
Furthermore, the overall quality of the scene's structure and texture improved (\textit{e.g.}, the fork and chips in the figure).

\begin{table}[t]
\centering
\caption{\textbf{Ablation study for varying division factor $d$.}}
\vspace{-2mm}
\resizebox{\linewidth}{!}{
\begin{tabular}{c|c|c|c|c|c}
    \toprule
    ~ & LPIPS $\downarrow$ & SSIM $\uparrow$ & PSNR $\uparrow$ & CD $\downarrow$ & F-score (0.05) $\uparrow$ \\
    \midrule
    $d=2$ & 0.251 & 0.598 & 19.8 & 0.0088 & 0.692 \\
    $d=4$ & \second 0.240 & \second 0.611 & \second 20.4 & \second 0.0086 & \second 0.694 \\
    $d=8$ & \first 0.237 & \first 0.615 & \first 20.5 & \first 0.0079 &  \first 0.699 \\
    \bottomrule
\end{tabular}
}
\label{tab:ablation1}
\end{table}

\begin{table}[t]
\centering
\vspace{-2mm}
\caption{\textbf{Ablation study on prior initialization and optimization}}
\vspace{-2mm}
\resizebox{\linewidth}{!}{
\begin{tabular}{l|c|c|c|c|c}
    \toprule
    ~ & LPIPS $\downarrow$ & SSIM $\uparrow$ & PSNR $\uparrow$ & CD $\downarrow$ & F-score (0.05) $\uparrow$\\
    \midrule
    p.w. flow only & 0.606 & 0.209 & 9.63 & 0.0348 & 0.261 \\
    + initialize & 0.425 & 0.312 & 13.0 & \second 0.0083 & 0.693 \\
    + SS optim. & \second 0.400 & \second 0.333 & \second 13.8 & \first 0.0078 & \first 0.708 \\
    + \SLat\ optim. & \first 0.240 & \first 0.611 & \first 20.4 & 0.0086 & \second 0.694 \\
    \bottomrule
\end{tabular}
}
\label{tab:ablation2}
\vspace{-5mm}
\end{table}

\vspace{2mm}
\noindent
In addition, as in \cref{sec:quantitative}, we quantitatively evaluated the generated scenes. 
\Cref{tab:ablation1} and \cref{tab:ablation2} show the effectiveness of our proposed methods, consistent with the qualitative observation that increasing $d$ and providing priors enhances the quality of the scenes. 
From the geometric results in \cref{tab:ablation2}, we find that initialization and SS optimization refine the geometry, whereas \SLat\ optimization sometimes degrades it. 
Since \SLat\ optimization usually enhances a 3D scene's texture, there is a trade-off between geometry and texture when optimizing the \SLat.
\Cref{tab:ablation3}, conducted without optimization and with $n_\mathrm{iter}=1$ as in the qualitative experiment, demonstrates that under-noising is the best choice of $t_\mathrm{start}$ and $t_\mathrm{noise}$ in 3D completion.

\begin{table}[t]
\centering
\caption{\textbf{Ablation study on under-noising.}}
\vspace{-2mm}
\resizebox{\linewidth}{!}{
\begin{tabular}{c|c|c|c|c|c}
    \toprule
    $t_\mathrm{noise} \ / \ t_\mathrm{start}$ & LPIPS $\downarrow$ & SSIM $\uparrow$ & PSNR $\uparrow$ & CD $\downarrow$ & F-score (0.05) $\uparrow$\\
    \midrule
    0.6 / 0.6 & 0.388 & 0.324 & 13.4 & 0.0081 & 0.657 \\
    0.8 / 0.8 & 0.550 & 0.216 & 9.91 & 0.0292 & 0.378 \\
    \midrule
    \makecell{0.8 / 0.6 \\ (over-noise)} & 0.622 & 0.219 & 9.79 & 0.0518 & 0.249 \\
    \midrule
    \makecell{0.6 / 0.8 \\ (\textbf{under-noise})} & \textbf{0.387} & \textbf{0.327} & \textbf{13.5} & \textbf{0.0078} & \textbf{0.680} \\
    \bottomrule
\end{tabular}
}
\label{tab:ablation3}
\vspace{-5mm}
\end{table}

\section{Conclusion}
\label{conclusion}

We propose a training-free 3D scene generation pipeline, Extend3D. 
By the extended latent space of the pre-trained object-centric model, we enabled scalable 3D scene generation. 
We demonstrate that our method (overlapping patch-wise flow, initialization, and optimization) and its schemes (iterative SDEdit, under-noising, and 3D-aware optimization objectives) achieve notable improvements in image-guided 3D scene generation. 

\vspace{2mm}
\noindent\textbf{Limitations.}
We found three limitations in our method. 
Firstly, occluded region completion is sometimes incomplete, for example, the one representing a room in \cref{More Results}. 
Secondly, \SLat\ optimization requires considerable memory, especially for large scenes (computational cost analysis is provided in \cref{cost}). 
Lastly, our framework shows limited performance on street-level images.
The problem is due to a significant mismatch between the scales of the $x$ and $y$ coordinates, arising from the vanishing points.
It would be a direction for future work to implement 3D generation from a wider range of image types.

\section*{Acknowledgements}
This work was supported by the IITP grant funded by MSIT [NO.RS-2021-II211343: AI Graduate School (Seoul National University) (5\%), NO.RS-2025-25442338: AI Star Fellowship (45\%), and NO.RS-2025-02303703: Real-world multi-space fusion and 6DoF free-viewpoint immersive visualization for extended reality (50\%)].

{
    \small
    \bibliographystyle{ieeenat_fullname}
    \bibliography{main}
}
\clearpage
\appendix
\maketitlesupplementary
\section{Appendix}

\subsection{Algorithms}
\label{Algorithms}
\begin{figure}[H]
\centering
\begin{minipage}[t]{0.49\textwidth}
\vspace{-2em}
\begin{algorithm}[H]
    \caption{Sparse Structure Generation}
    \label{alg:ss_algorithm}
    \begin{algorithmic}[1]
    \State \textbf{Input}: $\image$, $\pc$
    \\
    \State $\occ_0 \gets \mathbf{1}_\pc$
    \State Define schedule \\
        \quad $[t_\mathrm{start}=t_1>...>t_k=0]$
    \For{$0 \leq n < n_\mathrm{iter}$}
        \State $\elatent_0^{(g)} \gets \mathcal{E}(\occ_n)$
        \State Sample $\noise \sim \mathcal{N}(\bm0, \mI)$
        \State $\elatent_{t_1} \gets (1-t_\mathrm{noise})\cdot \elatent_0^{(g)}+t_\mathrm{noise}\cdot\noise$
        \For{$1\leq m<k$}
            \State Initialize $\oefvec \gets \efvec(\elatent_{t_m}, \image, t_m)$
            \State Optimize $\oefvec$ with Adam
            \State $\elatent_{t_{m+1}} \gets \elatent_{t_m}+(t_{m+1}-t_m)\cdot\oefvec$
        \EndFor
        \State $\occ_{n+1} \gets (\mathcal{D}(\elatent_0) > 0$)
    \EndFor\\
    \Return $\{\pos:(\occ_{n_\mathrm{iter}})_\pos >0 \}$
    \end{algorithmic}
\end{algorithm}
\vspace{-2em}
\end{minipage}
\hfill
\begin{minipage}[t]{0.49\textwidth}
\begin{algorithm}[H]
    \caption{Structured Latent Generation}
    \label{alg:slat_algorithm}
    \begin{algorithmic}[1]
    \State \textbf{Input}: $\{p_i\}, \image, \camera$
    \\
    \State Initialize $\elatent_1$ with $\{\pos_i\}$
    \State Define schedule $[1=t_1>...>t_k=0]$
    \For{$1\leq m<k$}
        \State Initialize $\oefvec \gets \efvec(\elatent_{t_m}, \image, t_m)$
        \State Optimize $\oefvec$ with Adam
        \State $\elatent_{t_{m+1}} \gets \elatent_{t_m}+(t_{m+1}-t_m)\cdot\oefvec$
    \EndFor\\
    \Return $\elatent_0$
    \end{algorithmic}
\end{algorithm}
\vspace{-2em}
\vfill
\begin{algorithm}[H]
    \caption{Extend3D}
    \label{alg:extend3d}
    \begin{algorithmic}[1]
    \State \textbf{Input}: $\image$
    \\
    \State $\pc, \camera \gets MoGe2(\image)$
    \State $\{\pos_i\} \gets SS(\image, \pc)$
    \State $\elatent \gets \SLat(\{\pos_i\}, \image, \camera)$
    \\\Return $(\mathcal{D}_\mathrm{GS}(\elatent),  {\mathcal{D}}_\mathrm{mesh}(\elatent))$
    \end{algorithmic}
\end{algorithm}

\end{minipage}
\end{figure}

\subsection{Image Patchification}
\label{Image Patchification}
We precisely patchify an input image using point cloud coordinates extracted from the monocular depth estimator. 
We can map a pixel $\image_{x, y}$ to a coordinate of the point cloud $q_{x, y}$ in the extended latent space. 
When $q_{x, y}\in\sliding_{i, j}$, the 3D area corresponding to the given pixel is in the patch $(i, j)$ since the structure was initialized with the depth estimator. 
We therefore define the image patch $\psi_{i, j}(\image)$ by collecting all $\image_{x, y}$ where corresponding $q_{x, y}\in\sliding_{i, j}$, setting the other pixels to be black, and cropping out black regions so that the image becomes square.

\subsection{Dilated Sampling}
\label{Dilated Sampling}
We follow the recipe for dilated sampling in \cite{accdiffusion2024}.
In this section, we assume that the unextended latent shape is $K\times K\times K$ for the sake of generalization. We divide the extended latent into $K\times K$ non-overlapping patches so that the size of each patch is $a\times b\times K$. We then randomly sample a pillar of 3D latent in each patch. The sampled pillars are attached by maintaining their relative positions to be a $K\times K \times K$ shaped latent. We sample $a\times b$ samples without replacement, and we call them dilated samples. The dilated samples are passed through the pre-trained model with the image condition $C_\image$ without image patchification. When dilated sampling is applied, \cref{extended vector} is altered to be:
\begin{equation}
\begin{aligned}
    \efvec(\elatent_t, \image, t)=~&
    (1-\gamma_t)
    \mathrm{PatchWise}(\elatent_t, C_\image,t)
    \\ &+ 
    \gamma_t\mathrm{Dilated}(\elatent_t, C_\image,t)
    ,
\end{aligned}
\end{equation}
\begin{equation}
    \gamma_t=0.5\cos^\alpha(\pi-\pi t)+0.5,
\end{equation}
where $\mathrm{PatchWise}$ is equal to \cref{extended vector} and $\alpha$ is a hyperparameter. We set $\alpha=5$ in our experiments. We use dilated sampling only for sparse structure generation because we empirically observed that it worsens texture when applied to structured latent generation.

\subsection{Computational Cost}
\label{cost}

\begin{table}[h]
\caption{\textbf{Computational costs of 3D generation methods.} \\ \centering{($a=b=2, d=4, n_\mathrm{iter}=5$)}}
\centering
\resizebox{0.7\linewidth}{!}{
\begin{tabular}{c|c|c}
    \toprule
     & VRAM (GB) & Time (min) \\
     \midrule
     Trellis~\citep{trellis2025} & 17 & 0.06 \\
     Hunyuan3D~\citep{hunyuan2} & 56 & 1.78 \\
     \midrule
     SynCity~\citep{syncity2025} & 49 & 52.0  \\
     EvoScene~\citep{evoscene} & 68 & 35.3 \\
     \textbf{Ours} & \textbf{28} & \textbf{14.1} \\
    \bottomrule
\end{tabular}
}
\label{tab:cost}
\end{table}

\noindent
We compared the computational cost of our method with previous methods in \cref{tab:cost}, using an NVIDIA RTX Pro 6000. Although our pipeline is heavier than object-centric methods (Trellis or Hunyuan3D), it requires substantially less memory and time than other scene-level pipelines.

\vspace{2mm}
\noindent
\textbf{Memory cost.}
For $a=b=2$ and $d=4$ case, the peak GPU memory used for our method is 28GB. Most of the memory is utilized in \SLat\ optimization process. Without \SLat\ optimization, it took 14GB VRAM. Also, for $a=b=6$ and $d=4$, without the \SLat\ optimization, the peak memory was 61 GB. \Cref{tab:quantitative} shows that our Extend3D achieves promising results even without \SLat\ optimization. Therefore, we can ensure reasonable quality with limited GPU resources.

\vspace{2mm}
\noindent
\textbf{Inference time.} The computational complexity of our method is $O(abd^2n_\mathrm{iter})$. A larger (larger $a$ and $b$), better detailed (larger $d$), and more complete (larger $n_\mathrm{iter}$) scene requires more inference time, so there is a trade-off between output quality and inference time.

\subsection{Rigorous Definition}
\label{math}
Here, we provide a detailed and rigorous definition of $\patch_{i,j}^\mathrm{SS}$, $\patch_{i,j}^\mathrm{\SLat}$, $(\patch_{i,j}^\mathrm{SS})^{-1}$, and $(\patch_{i,j}^\mathrm{\SLat})^{-1}$ for clarity. Firstly, we define patchification $\patch_{i,j}$ through sliding windows as:
\begin{equation}
    \sliding_{i, j}^K=\left[\frac{iK}{d},\frac{iK}{d}+K\right)\times\left[\frac{jK}{d},\frac{jK}{d}+K\right)\times[K],
\end{equation}
\begin{equation}
    \patch_{i,j}^\mathrm{SS}(\elatent_t^\mathrm{SS})=(\elatent_t^\mathrm{SS})_{\sliding_{i, j}^N}
    \in \mathbb{R}^{N\times N\times N},
\end{equation}
\begin{equation}
\label{slat_patch_sampling}
\begin{aligned}
\patch_{i,j}^\mathrm{\SLat}\!\left(\elatent_t^\mathrm{\SLat}\right)
&= \Biggl\{
    \Bigl(
      \pos - \Bigl(\tfrac{iM}{d},\, \tfrac{jM}{d},\, 0\Bigr),
      \lat_t
    \Bigr)
    :\\
&\qquad
    (\pos, \lat_t) \in \elatent_t^\mathrm{\SLat},\;
    \pos \in \sliding_{i, j}^M
  \Biggr\}.
\end{aligned}
\end{equation}
The subtraction in \cref{slat_patch_sampling} is for coordinate normalization into $[M]^3$. 
Secondly, the inversions of these processes are defined as:
\begin{equation}
    (\patch_{i,j}^\mathrm{SS})^{-1}(\mX)_{x,y,z}=\mathbf{1}_{(x,y,z)\in \sliding^N_{i, j}}\cdot \mX_{x-\frac{iN}{d},y-\frac{jN}{d}, z},
\end{equation}
\begin{equation}
\begin{aligned}
(\patch_{i,j}^\mathrm{\SLat})^{-1}(\mX)
&= \Biggl(
    \mX
    + \Bigl(
        \Bigl(\tfrac{iM}{d},\, \tfrac{jM}{d},\, 0\Bigr),
        \bm0
      \Bigr)
  \Biggr)\\
&\quad \cup\;
  \Bigl\{
    (\pos, \bm0)
    : \pos \in \{\pos_i\},\;
      \forall \lat~ (\pos, \lat) \notin \mX
  \Bigr\},
\end{aligned}
\end{equation}
setting the value of other positions to zero.

\subsection{Ablation on Iterative SDEdit}
\begin{table}[h]
\caption{\textbf{Ablation study on varying number of iterations $n_\mathrm{iter}$.}}
\label{tab:ablation_iter}
\centering
\resizebox{0.6\linewidth}{!}{
\begin{tabular}{c|c|c}
\toprule
    ~ & CD $\downarrow$ & F-score (0.05) $\uparrow$ \\
\midrule
    $n_\mathrm{iter}=0$ & \third 0.0079 & 0.647 \\
\midrule
    $n_\mathrm{iter}=1$ & \first 0.0075 & \second 0.681 \\
    $n_\mathrm{iter}=2$ & \second 0.0077 & \first 0.688 \\
    $n_\mathrm{iter}=3$ & 0.0082 & \third 0.677 \\
\bottomrule
\end{tabular}
}
\end{table}

\begin{figure}
    \centering
    \includegraphics[width=\linewidth]{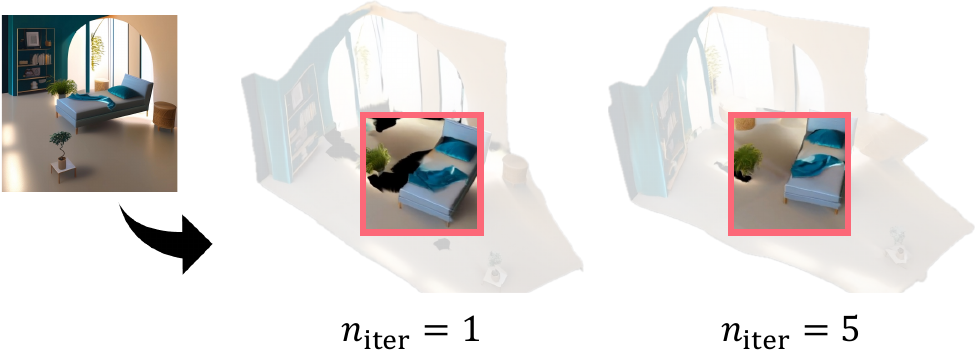}
    \caption{\textbf{Qualitative effect of iterative SDEdit}}
    \label{fig:n_iter}
\end{figure}

\noindent
We evaluated geometric results of different values of $n_\mathrm{iter}$. 
The result with $n_\mathrm{iter}=0$ represents the decoded \SLat\ with geometry from the monocular depth estimator. 
Compared to the $n_\mathrm{iter}=0$ case, with a single under-noised SDEdit step, the geometry improved noticeably, indicating that our method can address incompleteness in depth estimation. 
This also implies that our method is robust to the minor errors of the depth estimator. 
While a single SDEdit step improved results, additional SDEdit steps sometimes degraded output quality, as shown in \cref{tab:ablation_iter}.
Since the qualitative results, such as \Cref{fig:n_iter}, require iterations to complete the geometry, there is a trade-off between scene completion and geometric detail.

\subsection{More Results}
\label{More Results}
\begin{figure*}[p]
    \centering
    \includegraphics[width=\linewidth]{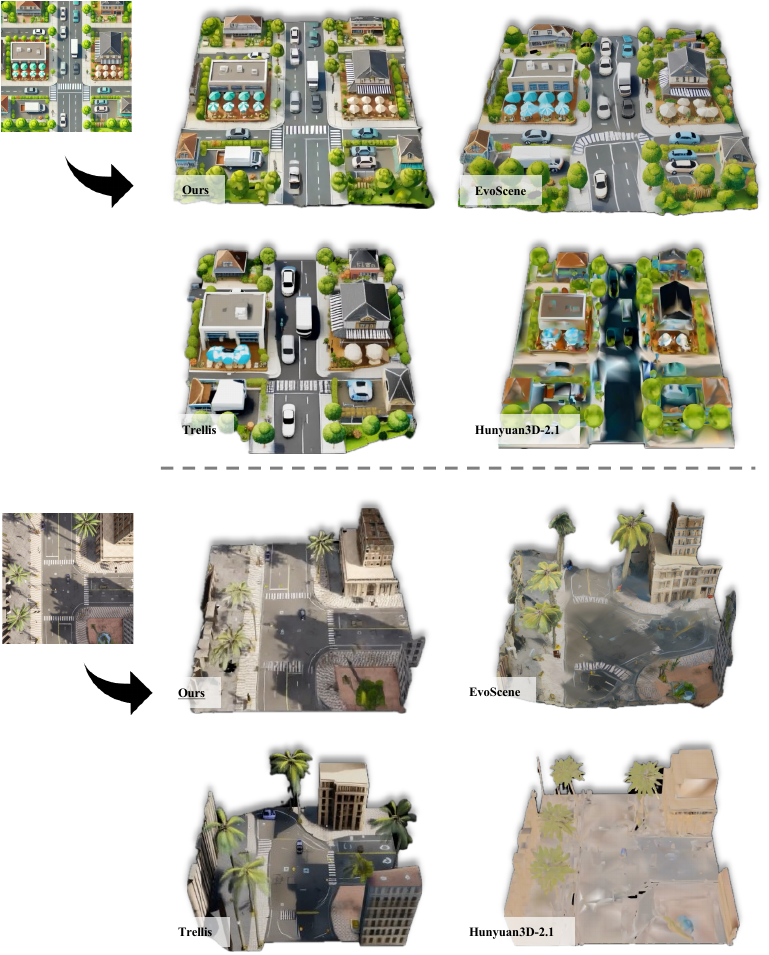}
    \caption{\textbf{Example results for diverse images.} In this figure, we set $a=b=2$. We compared our results with state-of-the-art open-source 3D generative models. The second image is from CarlaSC dataset~\citep{motionsc}. The other image is generated by ChatGPT~\citep{chatgpt}.}
    \label{fig:more1}
\end{figure*}

\begin{figure*}[p]
    \centering
    \includegraphics[width=0.95\linewidth]{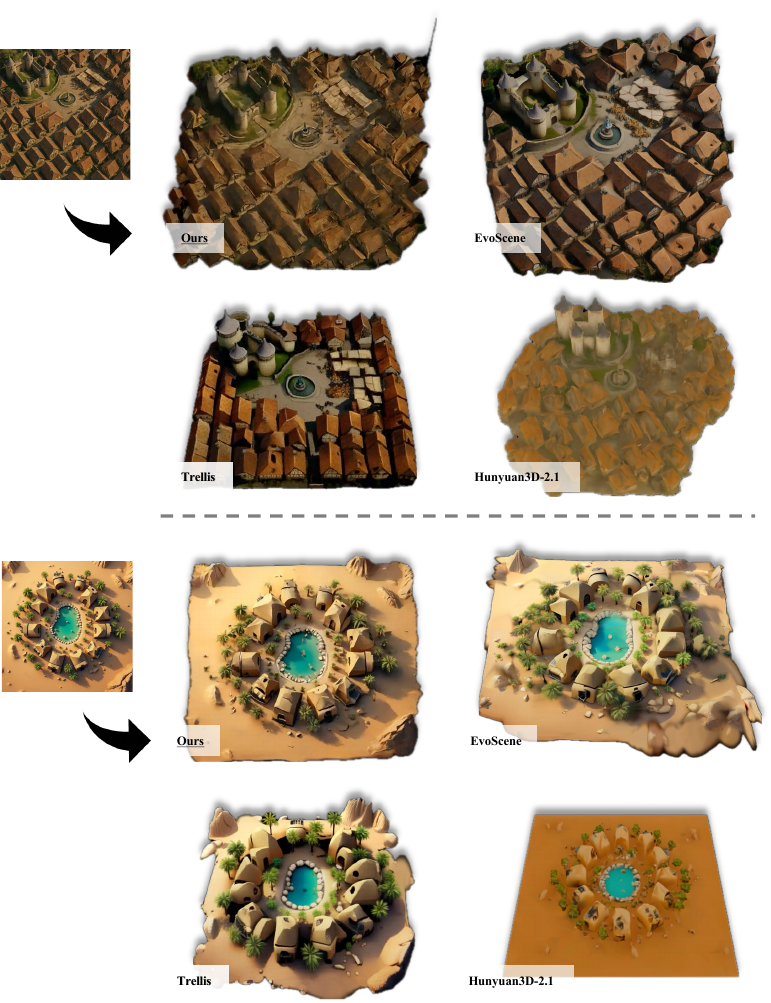}
    \caption{\textbf{Example results for diverse images.} In this figure, we set $a=b=2$. We compared our results with state-of-the-art open-source 3D generative models. The images are generated by ChatGPT~\citep{chatgpt} and Flux.1 [dev]~\citep{flux2024}.}
    \label{fig:more2}
\end{figure*}

\begin{figure*}[p]
    \centering
    \includegraphics[width=0.95\linewidth]{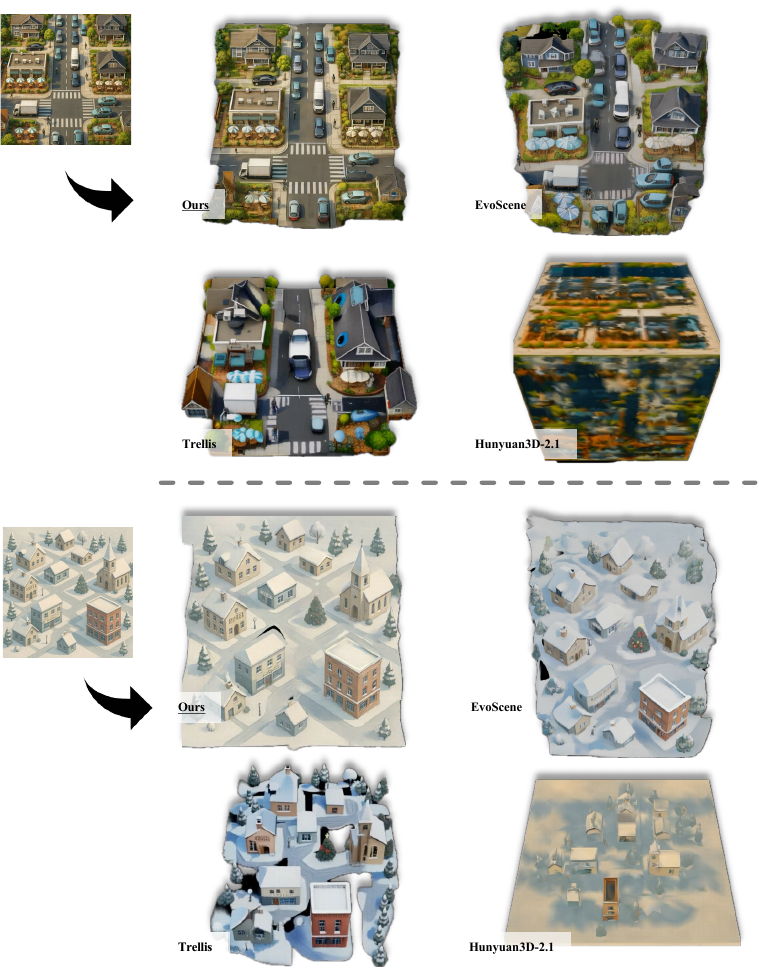}
    \caption{\textbf{Example results for diverse images.} In this figure, we set $a=b=2$. We compared our results with state-of-the-art open-source 3D generative models. The images are generated by ChatGPT~\citep{chatgpt}.}
    \label{fig:more3}
\end{figure*}

\begin{figure*}[p]
    \centering
    \includegraphics[width=\linewidth]{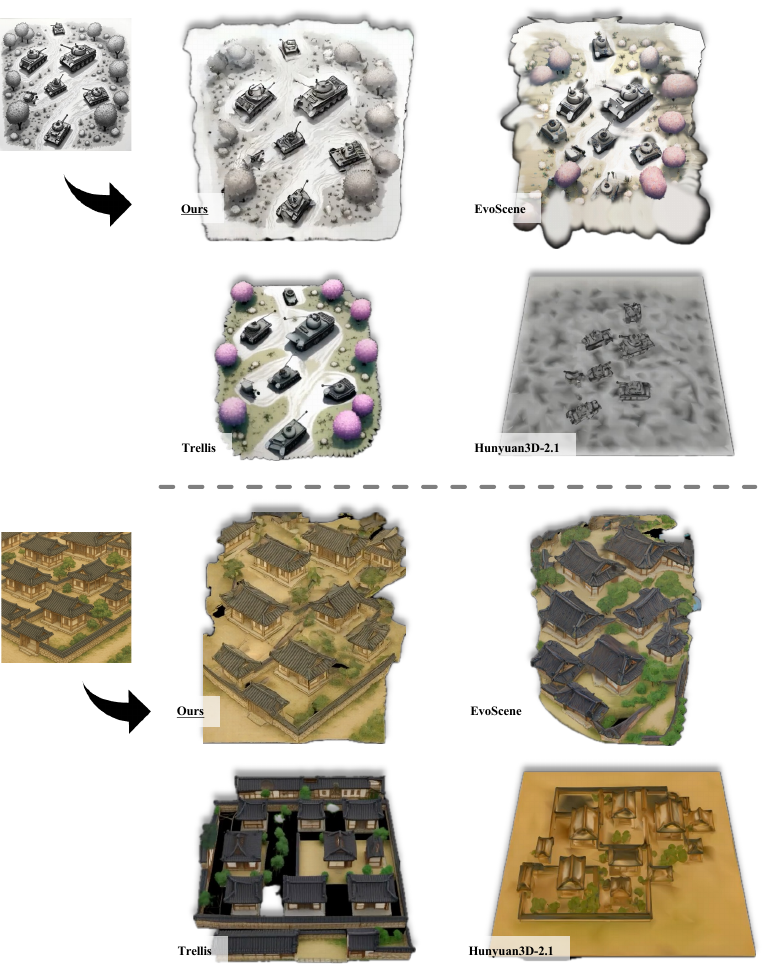}
    \caption{\textbf{Example results for diverse images.} In this figure, we set $a=b=2$. We compared our results with state-of-the-art open-source 3D generative models. The images are generated by ChatGPT~\citep{chatgpt}.}
    \label{fig:more4}
\end{figure*}

\begin{figure*}[p]
    \centering
    \includegraphics[width=\linewidth]{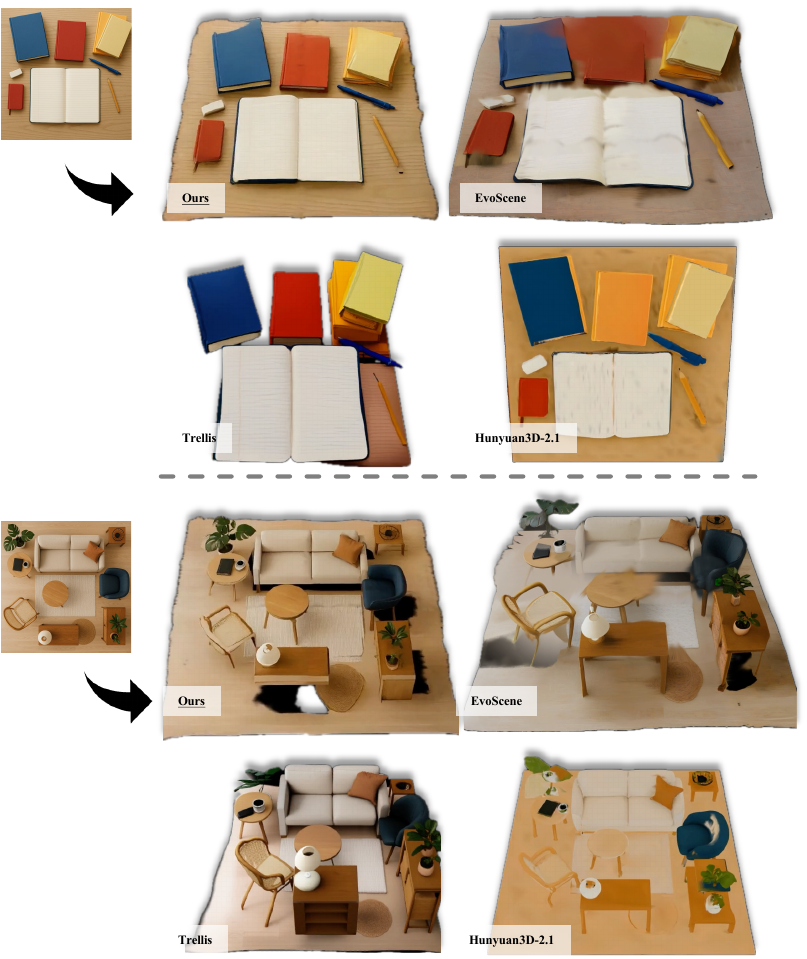}
    \caption{\textbf{Example results for diverse images.} In this figure, we set $a=b=2$. We compared our results with state-of-the-art open-source 3D generative models. The images are generated by ChatGPT~\citep{chatgpt}.}
    \label{fig:more5}
\end{figure*}

\begin{figure*}[p]
    \centering
    \includegraphics[width=\linewidth]{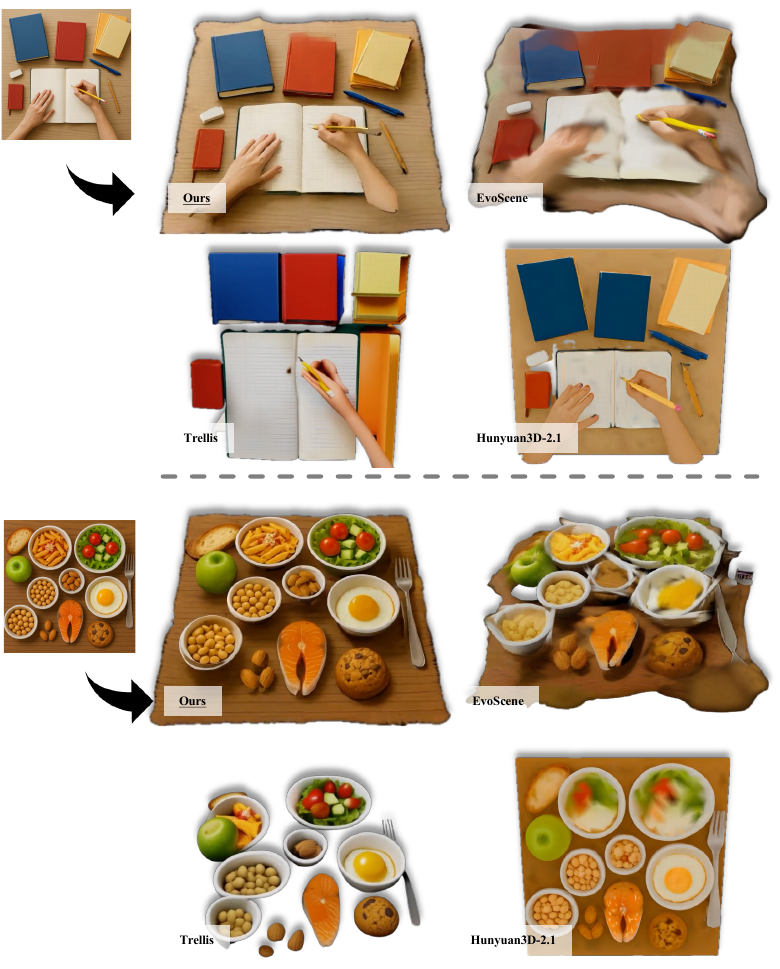}
    \caption{\textbf{Example results for diverse images.} In this figure, we set $a=b=2$. We compared our results with state-of-the-art open-source 3D generative models. The images are generated by ChatGPT~\citep{chatgpt}.}
    \label{fig:more6}
\end{figure*}

\begin{figure*}[p]
    \centering
    \includegraphics[width=\linewidth]{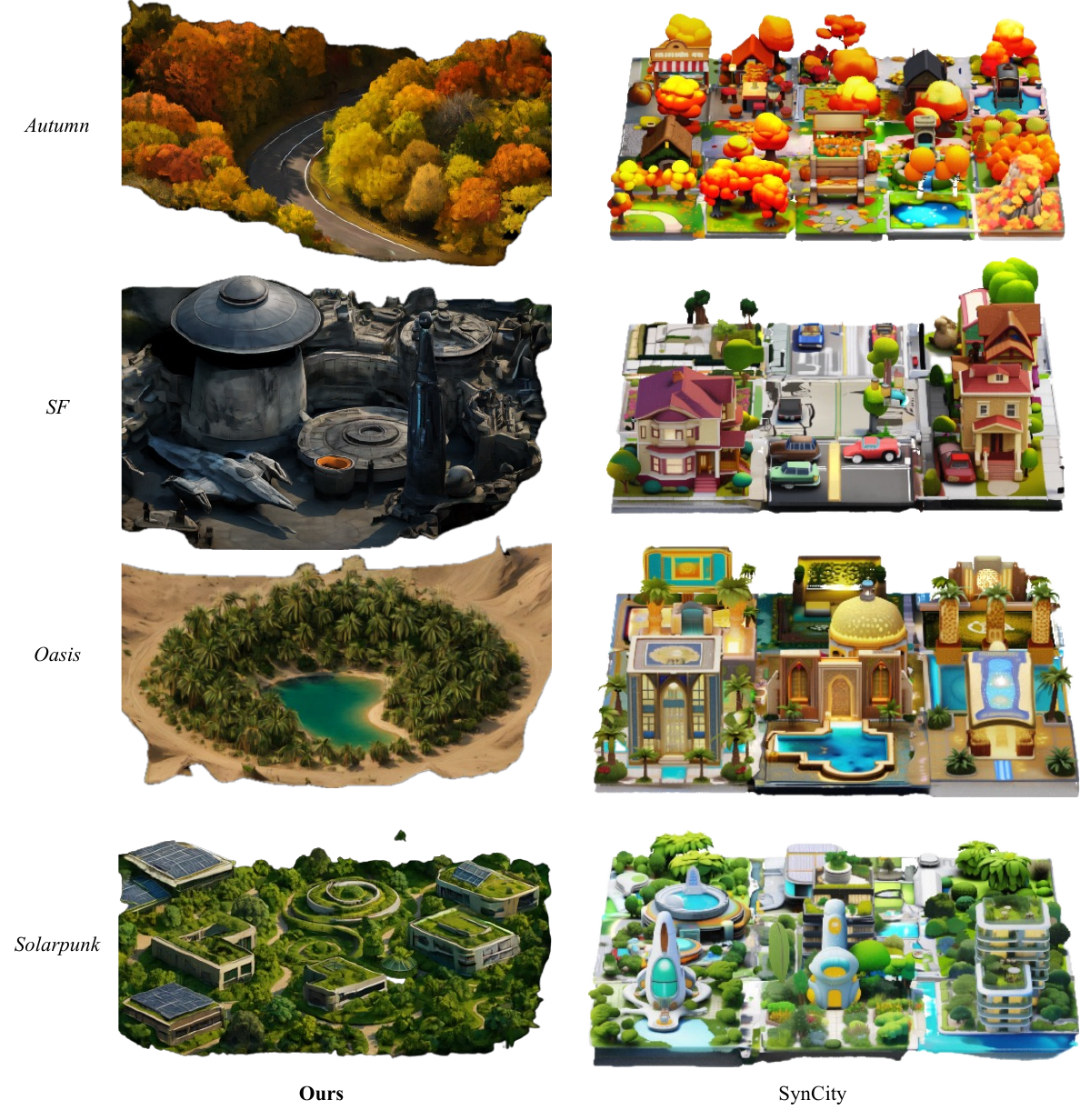}
    \caption{\textbf{Example results of comparison to SynCity.} We compared our results with SynCity~\citep{syncity2025}. Given a prompt, we first generated an image with ChatGPT~\citep{chatgpt} and used Extend3D for our results. The input prompts are on the left.}
    \label{fig:more8}
\end{figure*}

\begin{figure*}[h]
    \centering
    \includegraphics[width=\linewidth]{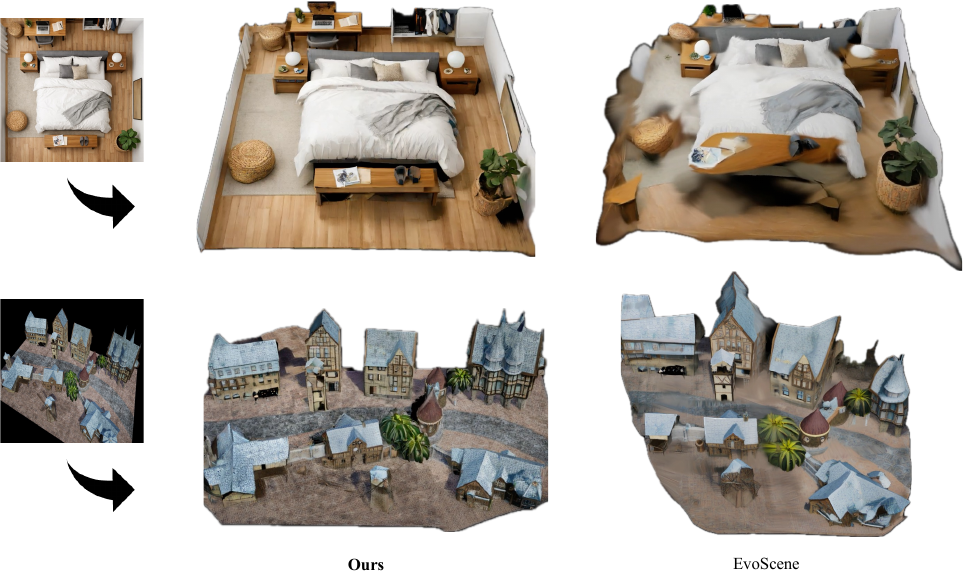}
    \caption{\textbf{Example results of comparison to EvoScene~\citep{evoscene}.} In this figure, we set $a$ = $b$ = 2. The images are from ChatGPT~\citep{chatgpt} and UrbanScene3D~\citep{urbanscene}.}
    \label{fig:evoscene}
\end{figure*}

\begin{figure*}[p]
    \centering
    \includegraphics[width=\linewidth]{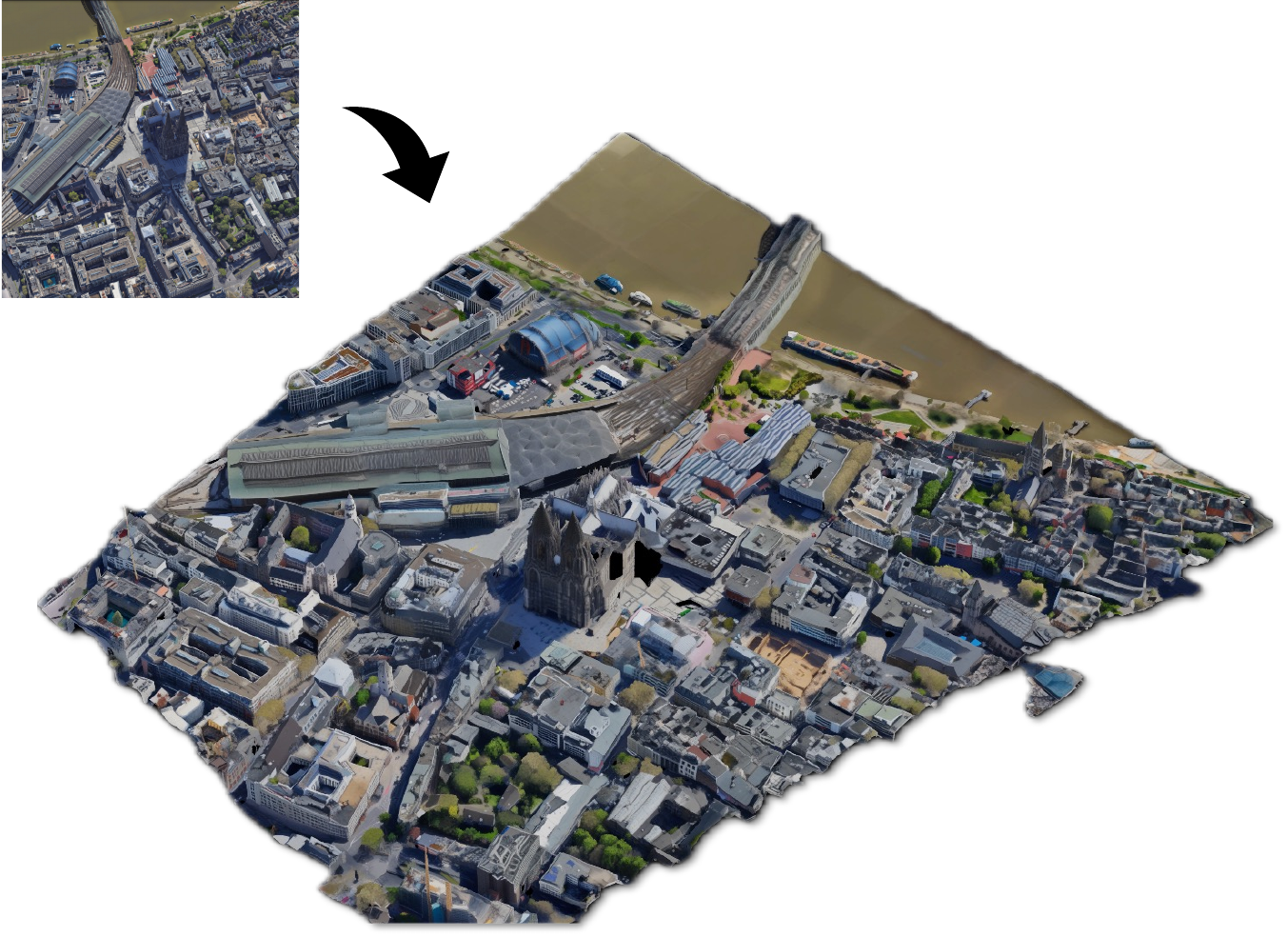}
    \caption{\textbf{The large scale result of Extend3D.} We generated large scale ($a=b=6$) 3D scene from the image of Köln captured from Google Earth~\citep{google-earth}. We didn't use \SLat\ optimization in this result due to the memory shortage.}
    \label{fig:large1}
\end{figure*}

\begin{figure*}[p]
    \centering
    \includegraphics[width=\linewidth]{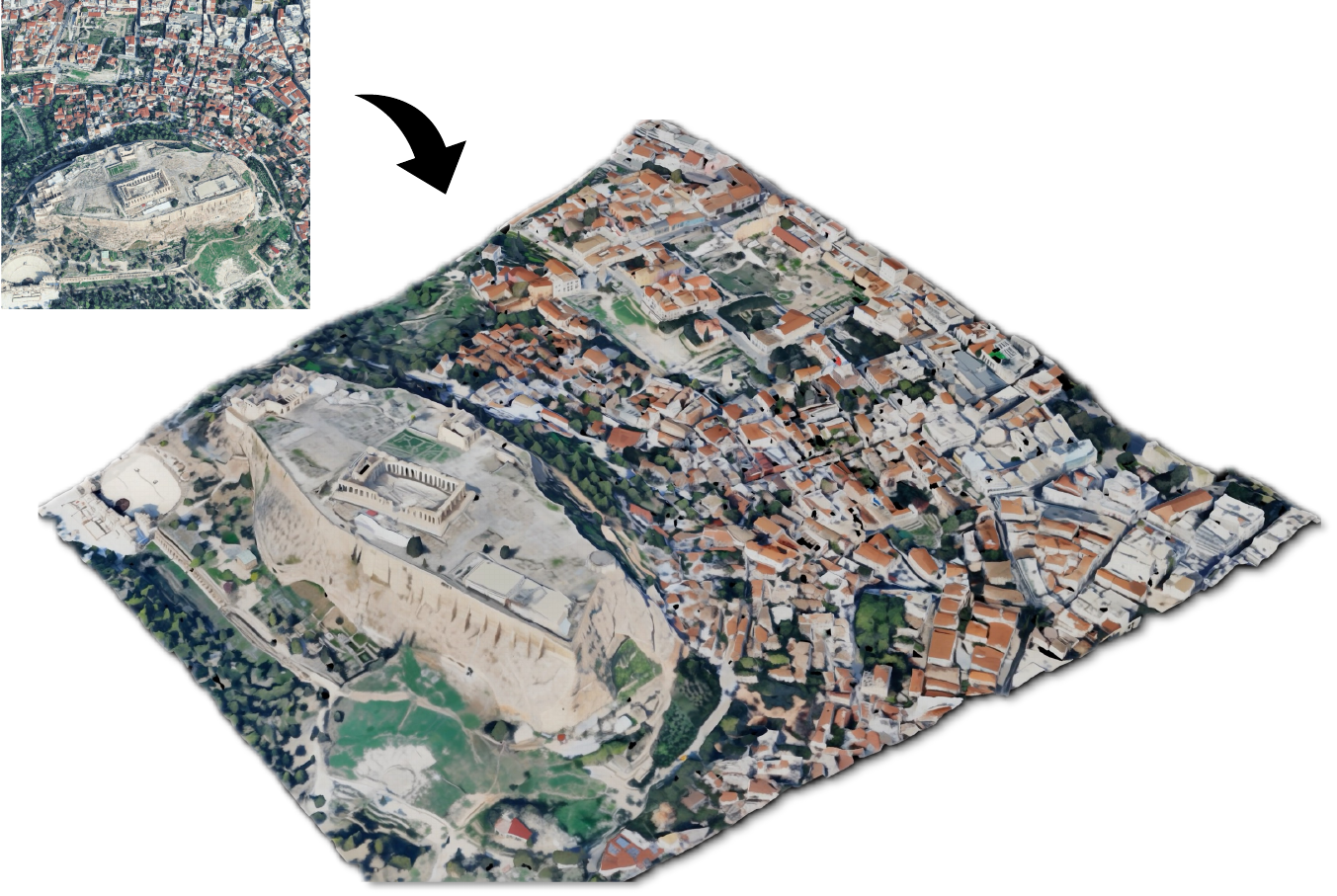}
    \caption{\textbf{The large scale result of Extend3D.} We generated large scale ($a=b=6$) 3D scene from the image of Athens captured from Google Earth~\citep{google-earth}. We didn't use \SLat\ optimization in this result due to the memory shortage.}
    \label{fig:large2}
\end{figure*}

We provide additional comparisons and large scene reconstruction results in \cref{fig:more1}, \cref{fig:more2}, \cref{fig:more3}, \cref{fig:more4}, \cref{fig:more5}, \cref{fig:more6}, \cref{fig:more8}, \cref{fig:evoscene}, \cref{fig:large1}, and \cref{fig:large2}.

\subsection{Quantitative Evaluation Set}
\begin{figure*}[p]
    \centering
    \includegraphics[width=\linewidth]{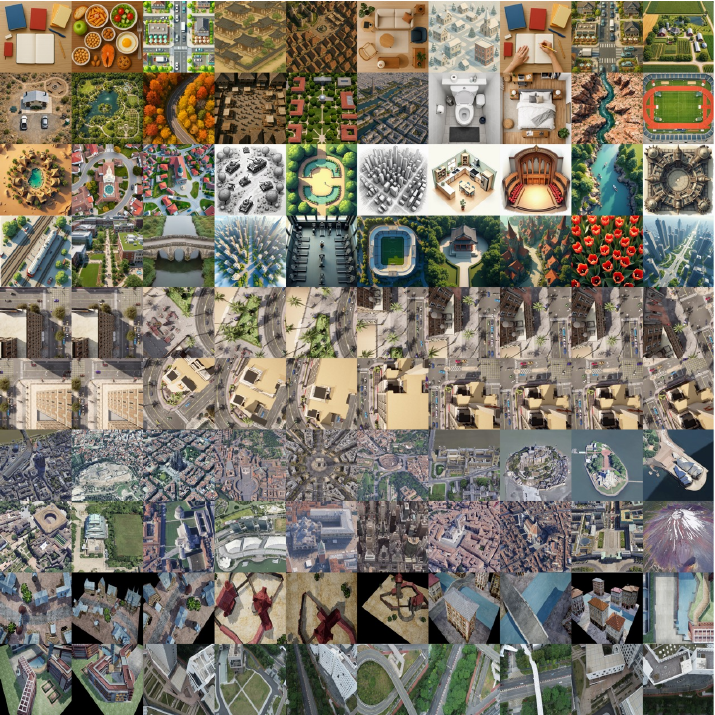}
    \caption{\textbf{100 images for appearance evaluation.} The images are from ChatGPT~\citep{chatgpt}, Flux.1 [dev]~\citep{flux2024}, CarlaSC~\citep{motionsc}, Google Earth~\citep{google-earth}, and UrbanScene3D~\citep{urbanscene}.}
    \label{fig:appearance_test}
\end{figure*}

\begin{figure*}
    \centering
    \includegraphics[width=\linewidth]{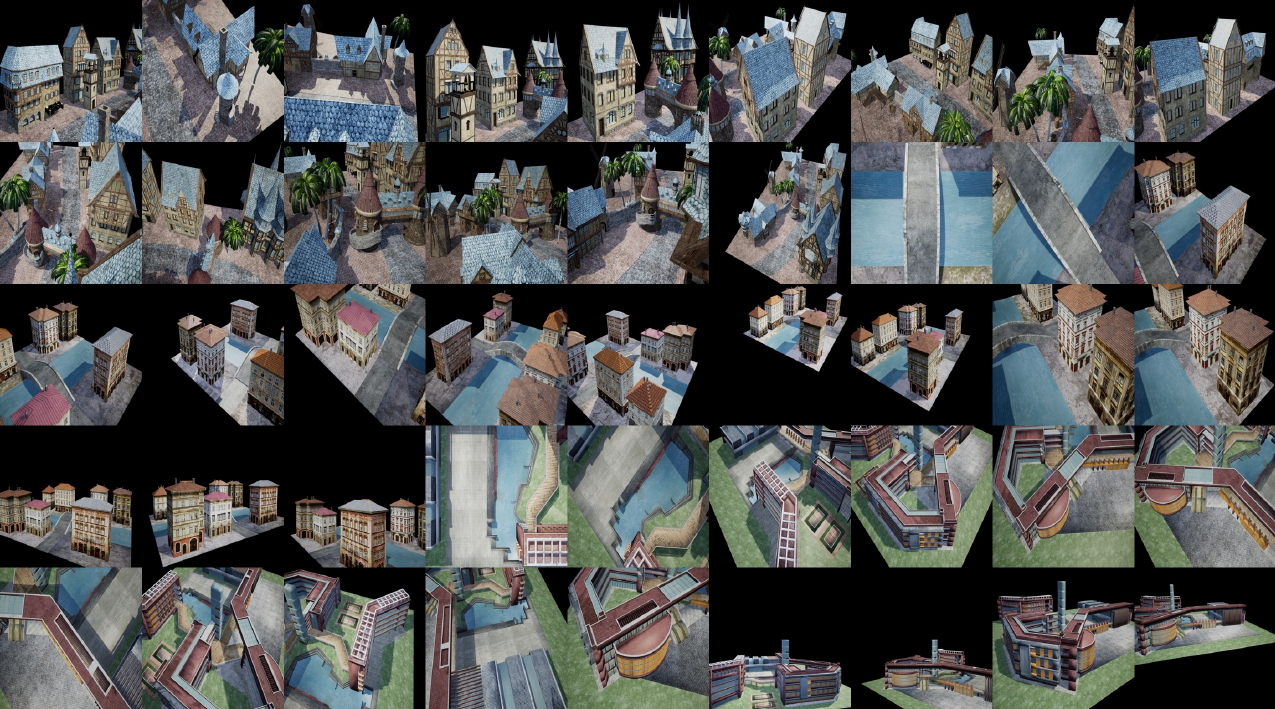}
    \caption{\textbf{45 images for geometry evaluation.} The images are from UrbanScene3D~\citep{urbanscene} synthetic scenes, together with ground truth meshes.}
    \label{fig:geometry_test}
\end{figure*}

The evaluation sets used for appearance evaluation (LPIPS, SSIM, PSNR) and for geometry evaluation (CD, F-score) are illustrated in \cref{fig:appearance_test} and \cref{fig:geometry_test}, respectively.
We used the keywords from the original SynCity research for a text-prompted generation comparison.
The keyword list is as follows: \textit{autumn}, \textit{campsite}, \textit{campus}, \textit{cyberpunk}, \textit{mars}, \textit{medieval market}, \textit{medieval}, \textit{oasis}, \textit{paris}, \textit{SF}, \textit{solarpunk}, and \textit{suburban}. Given a keyword, we used a prompt below for image generation before Extend3D:

\begin{quote}
\textit{Create a realistic, high-resolution aerial (bird’s-eye) view of \{keyword\}. The image should be perfectly framed so that no buildings, objects, or landscape features are cropped or cut off. Show the entire scene clearly from directly above, with natural lighting, realistic depth, and detailed textures. The result should look like a professional drone photograph taken at high altitude, with balanced composition and clean edges.}
\end{quote}

\subsection{\textbf{\SLat\ }Decoding}
While our main contribution is on the latent generation and most of our results in the paper are represented in 3D Gaussian Splatting, we also introduce an overlapping-patch SDF representation for mesh decoding. 
Unlike 3D Gaussians, if mesh patches are decoded disjointly and simply attached, the seams will be discontinuous, because there will be no faces. 
Therefore, similar to the overlapping patch-wise flow in our paper, we first decode \SLat\ into SDFs in an overlapping patch-by-patch manner.
These SDF patches are then combined into a single SDF by averaging the values in the overlapping regions. Inspired by TRELLISWorld~\citep{chen2025trellisworld}, we apply a cosine-based weighting to avoid near-zero SDF values at the seams.
Finally, we extract the mesh from the combined SDF in a single step.

Unlike meshes, we decode \SLat\ into 3D Gaussians with disjoint \SLat\ patches, since it is not necessary to consider connectivity in 3D Gaussians.

\newpage

\end{document}